\documentclass[12pt]{article}
\usepackage{amsmath}
\usepackage{mathrsfs}
\usepackage{amssymb}
\usepackage{times}
\usepackage{graphicx}
\usepackage{color}
\usepackage{multirow}

\usepackage{hyperref}

\usepackage{apacite}
\usepackage[authoryear]{natbib}
\usepackage{rotating}
\usepackage{bbm}
\usepackage{latexsym}
\usepackage{epstopdf}
\usepackage{graphicx}
\usepackage{caption}
\usepackage{subcaption}
\newtheorem{defn}{Definition}
\newtheorem{lema}{Lemma}
\newtheorem{thm}{Theorem}
\newtheorem{prop}{Proposition}
\newtheorem{remk}{Remark}

\newcommand{\captionfonts}{\normalsize}

\makeatletter
\long\def\@makecaption#1#2{%
  \vskip\abovecaptionskip
  \sbox\@tempboxa{{\captionfonts #1: #2}}%
  \ifdim \wd\@tempboxa >\hsize
    {\captionfonts #1: #2\par}
  \else
    \hbox to\hsize{\hfil\box\@tempboxa\hfil}%
  \fi
  \vskip\belowcaptionskip}
\makeatother

\begin{document}
\hspace{13.9cm}1

\ \vspace{20mm}\\

{\LARGE\center Dimensionality-Dependent Generalization Bounds for $k$-Dimensional Coding Schemes}

\ \\
{\bf \large Tongliang Liu$^{\displaystyle 1}$,~Dacheng Tao$^{\displaystyle 1}$,~Dong Xu$^{\displaystyle 2}$}\\
{$^{\displaystyle 1}$QCIS, University of Technology Sydney. tliang.liu@gamil.com; dacheng.tao@uts.edu.au}\\
{$^{\displaystyle 2}$School of Computer Engineering, Nanyang Technological University. dongxu@ntu.edu.sg}\\
%

{\bf Keywords:} Generalization bound, Bennett type inequality, covering number, $k$-dimensional coding schemes, non-negative matrix factorization, dictionary learning, sparse coding, $k$-means clustering and vector quantization.

\thispagestyle{empty}
\markboth{}{NC instructions}
\ \vspace{-0mm}\\
%
\begin{center} {\bf Abstract} \end{center}
The $k$-dimensional coding schemes refer to a collection of methods that attempt to represent data using a set of representative $k$-dimensional vectors, and include non-negative matrix factorization, dictionary learning, sparse coding, $k$-means clustering and vector quantization as special cases. Previous generalization bounds for the reconstruction error of the $k$-dimensional coding schemes are mainly \textcolor{black}{dimensionality-independent}. A major advantage of these bounds is that they can be used to analyze the generalization error when data is mapped into an infinite- or high-dimensional feature space. However, many applications use finite-dimensional data features. Can we obtain dimensionality-dependent generalization bounds for $k$-dimensional coding schemes that are tighter than dimensionality-independent bounds when data is in a finite-dimensional feature space? The answer is positive.
In this paper, we address this problem and derive a dimensionality-dependent generalization bound for $k$-dimensional coding schemes by bounding the covering number of the loss function class induced by the reconstruction error. The bound is of order $\mathcal{O}\left(\left(mk\ln(mkn)/n\right)^{\lambda_n}\right)$, where $m$ is the dimension of features, $k$ is the number of the columns in the linear implementation of coding schemes, $n$ is the size of sample, $\lambda_n>0.5$ when $n$ is finite and $\lambda_n=0.5$ when $n$ is infinite. We show that our bound can be tighter than previous results, because it avoids inducing the worst-case upper bound on $k$ of the loss function. The proposed generalization bound is also applied to some specific coding schemes to demonstrate that the dimensionality-dependent bound is an indispensable complement to the dimensionality-independent generalization bounds.


\section{Introduction}
The $k$-dimensional coding schemes \citep{KMaurerP10} are abstract and general descriptions of a collection of methods, all of which encode a data point $x\in\mathcal{H}$ as a representative vector $y\in\mathbb{R}^k$ by a linear map $T$, where $\mathcal{H}$ denotes the Hilbert space. These coding schemes can be formulated as follows:
\[\hat{y}=\arg\min_{y\in Y}\|x-Ty\|^2,\]
where $Y\subseteq \mathbb{R}^k$ is called the \emph{codebook} and the linear map $T\in\mathbb{R}^{m\times k}$ is called the \emph{implementation} of the codebook. The implementation projects the codebook back to the data source space. The dimension of a data point $x$ can be either finite or infinite. In this paper, we consider the data as having finite dimensions of features, that is $\mathcal{H}=\mathbb{R}^m$.

Each data point in $\mathcal{H}$ can be exactly or approximately reconstructed by a \emph{code} $y$ in the codebook. The \emph{reconstruction error} of a data point $x$ is defined as
\begin{eqnarray}\label{reconstructionerror}
f_T(x)=\min_{y\in Y}\|x-Ty\|^2.
\end{eqnarray}
The function $f_T(x)$, whose variables are $x$ and $T$, is also called the \emph{loss function}. Non-negative matrix factorization (NMF) \citep[see, e.g.,][]{lee99,fevotte2009nonnegative}, dictionary learning \citep[see, e.g.,][]{Donoho,Ivana}, sparse coding \citep[see, e.g.,][]{olshausen1996,amiri2014improved}, $k$-means clustering \citep[see, e.g.,][]{macqueen1967some,AnderbergCluster1973} and vector quantization \citep[see, e.g.,][]{gray1984vector,schneider2009adaptive} are specific forms of $k$-dimensional coding schemes, because they share the same form of the reconstruction error as equation (\ref{reconstructionerror}). They have achieved great successes in the fields of pattern recognition and machine learning for their superior performances on a broad spectrum of applications \citep[see, e.g.,][]{pehlevan2015hebbian,mairal2012task,hunt2012sparse,wright2009robust,schneider2009distance,dhillon2007weighted,quiroga2004unsupervised,kanungo2002efficient,abbott1999effect}.

Any coding scheme should find a proper implementation $T$. A natural choice for $T$ is the one that minimizes the \emph{expected reconstruction error}
\[R(T)=\int_{x}f_T(x)d\rho(x)=\int_{x}f_T(x)p(x)dx,\]
where $\rho(x)$ is a Borel measure of the data source, and $p(x)$ is the probability density function. However, in most cases, $p(x)$ is unknown, and $R(T)$ cannot be directly minimized. An alternative approach is the \emph{empirical risk minimization} (ERM) method \citep{vapnik2000,Cucker02onthe}. Given a finite number of independent and identically distributed observations $x_1,\ldots,x_n\in\mathbb{R}^m$, the \emph{empirical reconstruction error} with respect to $T$ is defined as
\[R_n(T)=\frac{1}{n}\sum_{i=1}^{n}f_T(x_i).\]
The ERM method searches for a $T_n$ that minimizes $R_n(T)$, and in the hope that $R(T_n)$ has a small distance to the expected reconstruction error $R(T^*)$, where
\[T^*=\arg\min_{T\in\mathcal{T}}R(T),\]
and $\mathcal{T}$ denotes a particular class of linear operators $T$.

A probabilistic bound on the defect
\[\sup_{T\in\mathcal{T}}\left|R(T)-R_n(T)\right|\]
is called the \emph{generalization (error) bound}. This paper focuses on this error bound in the framework of $k$-dimensional coding schemes. Although different restrictions are imposed on the choices of $\mathcal{T}$ and $Y$ for different concrete forms of $k$-dimensional coding schemes (for example, NMF requires both $\mathcal{T}$ and $Y$ to be non-negative, and sparse coding requires sparsity in $Y$), they are closely related. For example, \citet{Ding05onthe} showed that NMF with orthogonal $(y_1,\ldots,y_n )^\top$ is identical to $k$-means clustering of $\{x_1,\ldots,x_n\}$. Since these different forms of $k$-dimensional coding schemes are closely related, analyzing the generalization bounds together in this context has the advantages of exploiting the common properties and mutual cross-fertilization.

\subsection{Related work}
\citet{KMaurerP10} and \citet{gribonval2013sample} have performed the only known theoretical analyses on the generalization error in the framework of $k$-dimensional coding schemes. Other works have concentrated only on specific $k$-dimensional coding schemes. Since some previous works have studied \emph{consistency} performance, which considers the quantity
$R(T_n)-R(T^*)$
of the related ERM-based algorithms, we demonstrate the relationship between the generalization error and consistency performance here:
\begin{eqnarray*}
&&R(T_n)-R(T^*)\\
&&=R(T_n)-R_n(T_n)+R_n(T_n)-R_n(T^*)+R_n(T^*)-R(T^*)\\
&&\leq R(T_n)-R_n(T_n)+R_n(T^*)-R(T^*)\\
&&\leq 2\sup_{T\in\mathcal{T}}|R(T)-R_n(T)|.
\end{eqnarray*}
Thus, analyzing the generalization error provides an approach for analyzing the consistency performance, and the consistency performance provides directions to generalization error analysis. We review the generalization error and consistency performance of $k$-dimensional coding schemes together:
\begin{itemize}
  \item Non-negative matrix factorization (NMF). The only known generalization bounds of NMF are developed by \citet{KMaurerP10} and \citet{gribonval2013sample}.
  \item Dictionary learning. \citet{KMaurerP10} have developped dimensionality-independent generalization bounds. \citet{Vainsencher} and \citet{gribonval2013sample} have studied the dimensionality-dependent generalization bounds.
  \item Sparse coding. A generalization bound for sparse coding was first derived by \citet{KMaurerP10}, and subsequently extended by \citet{XuICML2012}, \citet{Gray2013}, \citet{sparsecodingICML}, and \citet{gribonval2013sample}. \citet{sparsecodingICML} derived a faster convergence rate upper bound of the consistency performance in a transfer learning setting.
  \item $K$-means clustering and vector quantization. Consistency performances of $k$-means clustering and vector quantization have mostly been studied for $\mathcal{H}=\mathbb{R}^m$. Asymptotic and non-asymptotic consistency performances have been considered by \citet{Pollard10}, \citet{Chou}, \citet{Linder}, \citet{Bartminmax}, \citet{Lindera}, \citet{Antosb}, \citet{Antos} and \citet{levrard2013fast}. Recently, \citet{biau}, \citet{KMaurerP10} and \citet{levrard2015nonasymptotic} developed dimensionality-independent generalization bounds for $k$-means clustering.
\end{itemize}
We are aware that these specific forms of $k$-dimensional coding schemes have many applications for finite-dimensional data, and only a few dimensionality-dependent methods have been developed to analyze the generalization bounds for all these coding schemes.

In this paper, we develop a dimensionality-dependent method to analyze the generalization bounds for the framework of $k$-dimensional coding schemes.
Our method is based on Hoeffding's inequality \citep{hoeffding1963probability} and the Bennett type inequalities \citep{boucheron2013concentration}, and directly bounds the covering number of the loss function class induced by the reconstruction error, which avoids inducing the worst-case upper bound on $k$ of the loss function.
Our method allows a generalization bound of order $\mathcal{O}\left(\left(mk\ln(mkn)/n\right)^{\gamma_n}\right)$, where $\gamma_n$ is much bigger than $0.5$ when $n$ is small, which delicately describes the non-asymptotic behavior of the learning process. However, when $n$ goes to infinity, $\gamma_n$ approaches to $0.5$. The obtained dimensionality-dependent generalization bound can be much tighter than the previous ones when the number $k$ of columns of the implementation is larger than the dimensionality $m$, which could often happen for dictionary learning, sparse coding, $k$-means clustering and vector quantization. We therefore obtain state-of-the-art generalization bounds for NMF, dictionary learning, sparse coding, $k$-means clustering and vector quantization.

The remainder of the paper is organized as follows.
We present our motivation in Section \ref{motivation} and main results in Section \ref{section2}.
In Section \ref{section3}, our results are applied to specific coding schemes and are empirically compared with state-of-the-art generalization bounds.
We prove our results in Section \ref{section5} and conclude the paper in Section \ref{section6}.

\section{Motivation}\label{motivation}
We first introduce the dimensionality-independent generalization bounds and demonstrate why our dimensionality-dependent bound complements them.

Assume that data points are drawn from a Hilbert space $\mathcal{H}$ with distribution $\mu$. For any $r\geq 0$, let $\mathcal{P}(r)$ denote the set of probability distributions on $\mathcal{H}$ supported on the closed ball of radius $r$ centered at the origin. In other words, $\mu\in\mathcal{P}(r)$ means that
 $P\{\|x\|\leq r\}=1.$
Let $\mathcal{T}$ be bounded in the operator norm, that is for every $T\in\mathcal{T}$, it holds that $\|Tv\|\leq c$ for all $v$ with $\|v\|\leq 1$. Then, we also have that the columns of $T$ are bounded as $\|Te_i\|\leq c, i=1,\ldots,k$, where $\{e_i|1\leq i\leq k\}$ is the orthonormal basis of $\mathbb{R}^k$.

The following two theorems are equivalent to the main theorems proved by \citet{KMaurerP10}, but are represented in a different way. They are dimensionality-independent generalization bounds obtained in the frame of the $k$-dimensional coding schemes. They exploited the Rademacher complexity technique \citep{BartlettRademacher02} which is suitable for deriving dimensionality-independent bounds \citep[see][]{biau}.
\begin{thm}\label{pontilone}
Assume that $\mu\in\mathcal{P}(r)$ and $Y$ is a closed subset of the unit ball of $\mathbb{R}^k$, and that there is $c\geq 0$ such that for all $T\in\mathcal{T}$, $\|Te_i\|\leq c,i=1,\ldots,k$. Suppose that the reconstruction error functions $f_T$ for $T\in\mathcal{T}$ have \textcolor{black}{a} range contained in $[0,b]$. For any $\delta\in(0,1)$, with probability at least $1-\delta$ in the independently observed data $x_1,\ldots,x_n\sim \mu$, we have
\begin{eqnarray*}
&&\sup_{T\in\mathcal{T}} \left|R(T)-R_n(T)\right|\leq (4crk+2c^2k^2)\sqrt{\frac{\pi}{n}}+b\sqrt{\frac{8\ln{2/\delta}}{n}}.
\end{eqnarray*}
\end{thm}

\begin{remk}
The dimensionality-independent generalization bound in Theorem \ref{pontilone} is valuable because it shows a \textcolor{black}{convergence rate of order $O(\sqrt{1/n})$}.
\end{remk}

\begin{thm}\label{pontiltwo}
Assume that $\mu\in\mathcal{P}(r)$ and $\|\mathcal{T}\|_Y=\sup_{T\in\mathcal{T}}\sup_{y\in Y}\|Ty\|$, and that the reconstruction error functions $f_T$ for $T\in\mathcal{T}$ have \textcolor{black}{a} range contained in $[0,b]$. For any $\delta\in(0,1)$, with probability at least $1-\delta$ in the independently observed data $x_1,\ldots,x_n\sim \mu$, we have
\begin{eqnarray*}
&&\sup_{T\in\mathcal{T}}\left|R(T)-R_n(T)\right|\\
&&\leq b\sqrt{\frac{\ln{2/\delta}}{2n}}+\frac{bk}{2}\sqrt{\frac{\ln\left(16n\|\mathcal{T}\|_Y^2\right)}{n}}+\frac{4+4\|\mathcal{T}\|_Y+\sqrt{8\pi}rk\|\mathcal{T}\|_Y}{\sqrt{n}}.
\end{eqnarray*}
If $\mathcal{H}$ is finite dimensional, the above result will be improved to
\begin{eqnarray*}
&&\sup_{T\in\mathcal{T}}\left|R(T)-R_n(T)\right|\\
&&\leq  b\sqrt{\frac{\ln{2/\delta}}{2n}}+\frac{b}{2}\sqrt{\frac{mk\ln{(16n\|\mathcal{T}\|_Y^2)}}{n}} +\frac{4+4\|\mathcal{T}\|_Y+\sqrt{8\pi}rk\|\mathcal{T}\|_Y}{\sqrt{n}}.
\end{eqnarray*}
\end{thm}

\begin{remk}
The condition that $Y$ is a closed subset of the unit ball of $\mathbb{R}^k$ can be easily achieved by controlling the \textcolor{black}{upper} bound of columns of $T$, because there is a trade-off between the bounds of columns of $T$ and the entries of $y\in Y$.
\end{remk}

\begin{remk}
We note that Theorems \ref{pontilone} and \ref{pontiltwo} are more complicated than the original results presented in \citep{KMaurerP10}. This is because we have removed the restrictions that $c\geq1$ and $\|\mathcal{T}\|_Y\geq 1$, which are required to simplify their results, to reveal the intrinsic relationships between the order of $k$ and the Rademacher complexities (discussed below). The proof methods of Theorems \ref{pontilone} and \ref{pontiltwo} in this paper are exactly the same as those \textcolor{black}{presented} by \citet{KMaurerP10}.
\end{remk}

We note that if $y$ is in the unit ball of $\mathbb{R}^k$, then
\begin{eqnarray*}
&f_T(x)&=\min_{y\in\mathbb{R}^{k\times 1}}\|x-Ty\|^2\leq\min_{y\in\mathbb{R}^{k\times 1}}\left(\|x\|^2+\|Ty\|^2\right)\leq r^2+\min_{y\in\mathbb{R}^{k\times 1}}\| Ty\|^2\\
&&= r^2+\min_{y\in\mathbb{R}^{k\times 1}}\sum\limits_{i,j}^{k}\left<y_iTe_i,y_jTe_j\right>\leq r^2+\min_{y\in\mathbb{R}^{k\times 1}}\sum\limits_{i,j}^{k}\|y_iTe_i\|\|y_jTe_j\|\\
&&\leq r^2+c^2k^2,
\end{eqnarray*}
where $r$, the upper bound of the data point, can be reduced by normalization. However, $k$ is a fixed integer, whose value is usually large in practice. Thus, $c^2k^2$ is \textcolor{black}{the} dominant factor in the upper bound of $f_T$. It is evident that $f_T$ has the worst-case upper bound on $k$ of order $\mathcal{O}(k^2)$, \emph{i.e.,} the dependency \emph{w.r.t.} $k$ of the upper bound of $f_T$ has the worst case order $\mathcal{O}(k^2)$. However, for some special forms of $k$-dimensional coding schemes, the upper bound of $f_T$ has a very small order about $k$. Taking NMF as an example, the order about $k$ is zero because
\begin{eqnarray*}
&&f_T(x)=\min_{y\in\mathbb{R}_+^k}\|x-Ty\|^2\leq\|x\|^2+\|T0\|^2\leq r^2.
\end{eqnarray*}
It is evident that the term $2c^2k^2\sqrt{\pi/n}$ in Theorem \ref{pontilone} has the same order as that of the worst-case upper bound on $k$ of $f_T$. It will therefore be loose for some specific $k$-dimensional coding schemes. \citet{KMaurerP10} introduced the proof method of Theorem \ref{pontiltwo} to overcome this problem; however, the term $rk\|\mathcal{T}\|_Y\sqrt{8\pi/n}$ implies that the problem is only partially solved, because $rk$ represents the worst-case upper bound on $k$ of $\sqrt{f_T}$ (details can be found in the proof therein). For example, in NMF, the term $rk\|\mathcal{T}\|_Y\sqrt{8\pi/n}$ is of order $\mathcal{O}(\sqrt{k^3/n})$ (discussed below in Remark \ref{remk4}). The dimensionality-dependent bound in Theorem \ref{pontiltwo} faces the same problem because the proof method computes the Rademacher complexity, corresponding to which part the obtained bound is dimensionality-independent and involves the worst-case upper bound on $k$ of $\sqrt{f_T}$.

We try to avoid the aforementioned worst case by employing a covering number method to measure the complexity of the induced loss function class $F_\mathcal{T}=\{f_T|T\in\mathcal{T}\}$. However, in our setting, the dimensionality $m$ of data space must be finite.

\section{Main results}\label{section2}
Before presenting our main results,
we first introduce the definition of \emph{covering number} $\mathcal{N}_p(F,\epsilon,n)$ \citep{Zhang02coveringnumber}.
\begin{defn}\label{coveringnumber}
Let $\mathcal{B}$ be a metric space with metric $d$. Given observations $X=\{x_1,\ldots,x_n\}$, and vectors $f(X)=\{f(x_1),\ldots,f(x_n)\}\in \mathcal{B}^n$, the covering number in $p$-norm, denoted as $\mathcal{N}_p(F,\xi,X)$, is the minimum number $m$ of a collection of vectors $v_1,\ldots, v_m\in \mathcal{B}^n$, such that $\forall f\in F,\exists v_j$:
\begin{eqnarray*}
&&\|d(f(X),v_j)\|_p=\left[\sum\limits_{i=1}^{n}d(f(x_i),v_j^i)^p\right]^{1/p}\leq n^{1/p}\xi,
\end{eqnarray*}
where $v_j^i$ is the $i$-th component of vector $v_j$. We also define $\mathcal{N}_p(F,\xi,n)=\sup_{X}\mathcal{N}_p(F,\xi,X)$.
\end{defn}

Let $\mathcal{T}=\mathbb{R}^{m\times k}$. We can upper bound the covering number of the induced loss function class of any $k$-dimensional coding scheme.

\begin{lema}\label{l1}
Let $F_\mathcal{T}=\{f_T|T\in\mathcal{T}, \mathcal{T}=\mathbb{R}^{m\times k}\}$ be the loss function class induced by the reconstruction error for a $k$-dimensional coding scheme. We have
\[\ln\mathcal{N}_1(F_\mathcal{T},\xi',n)\leq mk\ln\left(\frac{4(r+ck)\sqrt{m}ck}{\xi'}\right).\]
\end{lema}

By employing Hoeffding's inequality \citep{hoeffding1963probability}, we can derive a dimensionality-dependent generalization bound for $k$-dimensional coding schemes.

\begin{thm}[main result one]\label{mainone}
Assume that $\mu\in\mathcal{P}(r)$ and $Y$ is a closed subset of the unit ball of $\mathbb{R}^k$, and that there is $c\geq 0$ such that for all $T\in\mathcal{T}$, $\|Te_i\|\leq c, i=1,\ldots,k$, and that the functions $f_T$ for $T\in\mathcal{T}$ have \textcolor{black}{a} range contained in $[0,b]$. For any $\delta\in(0,1)$, with probability at least $1-\delta$, we have
\begin{eqnarray*}
&&\sup_{T\in\mathcal{T}}|R(T)-R_n(T)|\leq \frac{2}{n} +b\sqrt{\frac{mk\ln\left(4(r+ck)\sqrt{m}ckn\right)+\ln{2/\delta}}{2n}}.
\end{eqnarray*}
\end{thm}

Our result is dimensionality-dependent. Compared to the bound in Theorem \ref{pontiltwo}, our bound could be tighter if $m\ln{m}\leq k\|\mathcal{T}\|_Y^2$.

\begin{remk}\label{remk4}
Let us take NMF for example to show how our method avoids inducing the worst-case upper bound on $k$ of the loss function compared to those of Theorems \ref{pontilone} and \ref{pontiltwo}. Regarding NMF,
\begin{eqnarray*}
&&\|\mathcal{T}\|_Y=\sup_{T\in\mathcal{T}}\sup_{y\in Y}\|Ty\|= \sup_{T\in\mathcal{T}}\sup_{y\in Y}\left\|\sum_{i=1}^{k}y_iTe_i\right\|=\sup_{y\in Y}c\sum_{i=1}^{k}\left\|y_ie_i\right\|= c\sqrt{k}.
\end{eqnarray*}
If we only consider the order of $m, k$ and $n$, our bound is of order $\mathcal{O}(\sqrt{km\ln{(mkn)}/n})$ while Theorem \ref{pontilone} has order $\mathcal{O}(\sqrt{k^4/n})$ and Theorem \ref{pontiltwo} is of order $\mathcal{O}(\sqrt{k^3/n}+\sqrt{k^2\ln(kn)/n})$. Our bound is tighter when $m\ln{m}\leq k^2$.
\end{remk}
\begin{remk}
For dictionary learning, sparse coding, $k$-means clustering and vector quantization, the number $k$ of the columns of the linear implementation may be larger than the dimensionality $m$. If $k>m$, our bound will be much tighter than the dimensionality-independent generalization bound.
\end{remk}

\begin{remk}
According to the proofs of Lemma \ref{l1} and Theorem \ref{mainone}, our result is based on the estimation of the Lipschitz constant of the loss function $f_T(x)$ w.r.t. the implementation $T$. Particularly, we proved the property $|f_T(x)-f_{T'}(x)|\leq L |T-T'|$ for all $T$ and $T'$ in $\mathcal{T}$, where $L$ is a constant depending on a specific $k$-dimensional coding scheme. Similar to our idea, \citet{gribonval2013sample} also developed dimensionality-dependent generalization bounds for $k$-dimensional coding schemes. However, their method is different from ours. Their results are essentially based on the property that $|f_T(x)-f_{T'}(x)|\leq L' \|T-T'\|_{1\rightarrow 2}$ for all $T$ and $T'$ in $\mathcal{T}$, where $L'$ is also a constant and the operator norm $\|\cdot\|_{1\rightarrow 2}$ of an $m\times k$ matrix $A=[A_1,\ldots,A_k]$ is defined as $\|A\|_{1\rightarrow 2}=\sup_{\|\alpha\|_1\leq 1}\|A\alpha\|_2$. As a result, under some assumptions (see assumptions A1-A4, B1-B3 and C1-C2 therein) and with high probability, they have that
\begin{eqnarray*}
&\sup_{T\in\mathcal{T}}|R(T)-R_n(T)|&\leq 3c\sqrt{\frac{mk\cdot\max(\ln\frac{2L'C}{c},1)\ln n}{n}}\\
&&+ c\sqrt{\frac{mk\cdot\max(\ln\frac{2L'C}{c},1)+\ln 2/\delta}{n}},
\end{eqnarray*}
where $c, C, T$ are constants depending on a specific $k$-dimensional coding scheme. Note that in most applications, $\ln\frac{2L'C}{c}>1$ and $\ln n>1$. Their bound could be looser than the derived bound in Theorem \ref{mainone} because in the cases, it holds that $\ln\frac{2L'C}{c}\ln n>\ln\frac{2L'C}{c}+\ln n$. Detailed comparisons are presented in Section \ref{section3}.
\end{remk}

The result in Theorem \ref{mainone} can be improved by exploiting Bennett type inequalities.
We can make the upper bound to have either a smaller constant or a faster convergence rate as follows.

By employing Bernstein's inequality, we show that a tighter generalization bound of $k$-dimensional coding schemes than that in Theorem \ref{mainone} can be derived.
\begin{thm}[main result two]\label{mainthree}
Assume that $\mu\in\mathcal{P}(r)$ and $Y$ is a closed subset of the unit ball of $\mathbb{R}^k$, and that there is $c\geq 0$ such that for all $T\in\mathcal{T}$, $\|Te_i\|\leq c, i=1,\ldots,k$, and that the functions $f_T$ for $T\in\mathcal{T}$ have \textcolor{black}{a} range contained in $[0,1]$. For any $\delta\in(0,1)$, with probability at least $1-\delta$, we have
\begin{eqnarray*}
&\sup_{T\in\mathcal{T}}|R(T)-R_n(T)|&\leq \frac{2}{n} +\frac{5\left(mk\ln\left(4(r+ck)\sqrt{m}ckn\right)+\ln{2/\delta}\right)}{n}\\
&&+\sqrt{\frac{2R_n(T)\left(mk\ln\left(4(r+ck)\sqrt{m}ckn\right)+\ln{2/\delta}\right)}{n}}.
\end{eqnarray*}
\end{thm}


\begin{remk}
The upper bound in Theorem \ref{mainthree} can be much tighter than that in Theorem \ref{mainone}. The dominant term in the upper bound of Theorem \ref{mainthree} is $\sqrt{\frac{2R_n(T)\left(mk\ln\left(4(r+ck)\sqrt{m}ckn\right)+\ln{2/\delta}\right)}{n}}$. Since the empirical reconstruction error $R_n(T)$ is no bigger and sometimes much smaller than $1$, the upper bound in Theorem \ref{mainthree} can therefore be much tighter than that in Theorem \ref{mainone}.
\end{remk}

We can represent the result by using the inequlaity that for all $a,b,\lambda>0$, $\sqrt{2ab}<\lambda a+\lambda^{-1}b/4$.
\begin{prop}\label{mainthreenew}
Assume that $\mu\in\mathcal{P}(r)$ and $Y$ is a closed subset of the unit ball of $\mathbb{R}^k$, and that there is $c\geq 0$ such that for all $T\in\mathcal{T}$, $\|Te_i\|\leq c, i=1,\ldots,k$, and that the functions $f_T$ for $T\in\mathcal{T}$ have \textcolor{black}{a} range contained in $[0,1]$. For any $T\in\mathcal{T}$, any $\lambda>0$ and any $\delta\in(0,1)$, with probability at least $1-\delta$, we have
\begin{eqnarray*}
&&R(T)\leq (1+\lambda)R_n(T)+ \frac{2}{n}+\left(\frac{1}{4\lambda}+5\right)\frac{\left(mk\ln\left(4(r+ck)\sqrt{m}ckn\right)+\ln{2/\delta}\right)}{n}.
\end{eqnarray*}
\end{prop}

We have claimed that Theorem \ref{mainthree} and Proposition \ref{mainthreenew} can be tighter than Theorem \ref{mainone} by saying that $R_n(T)$ can be very small. However, sometimes, such a term could be large. If $R_n(T)>1/4$ (note that the reconstruction error function $f_T\in[0,1]$), Theorem \ref{mainthree} and Proposition \ref{mainthreenew} will be looser than Theorem \ref{mainone}.

The following theorem implies that by employing Bennett's type inequality, the generalization bound can be improved no matter what the value of $R_n(T)$ is. 
\begin{thm}[main result three]\label{mainfour}
Assume that $\mu\in\mathcal{P}(r)$ and $Y$ is a closed subset of the unit ball of $\mathbb{R}^k$, and that there is $c\geq 0$ such that for all $T\in\mathcal{T}$, $\|Te_i\|\leq c, i=1,\ldots,k$, and that the functions $f_T$ for $T\in\mathcal{T}$ have \textcolor{black}{a} range contained in $[0,1]$. For any $\delta\in(0,1)$, with probability at least $1-\delta$ it holds for all $T\in\mathcal{T}$ that
\begin{eqnarray*}
&&|R(T)-R_n(T)|\leq\frac{2}{n} +\left(\frac{mk\ln\left(4(r+ck)\sqrt{m}ckn\right)+\ln{\frac{2}{\delta}}}{\beta n}\right)^{\frac{1}{2-\frac{\ln\left(8\beta V/3\right)}{\ln|R(T)-R_n(T)|}}},
\end{eqnarray*}
when $V$ satisfies that $|R(T)-R_n(T)|\leq V\leq 3/8\beta$ and $\beta$ is any positive constant.
\end{thm}

\begin{remk}
Since $f_T(x)\leq 1$ in Theorem \ref{mainfour}, we have that  \textcolor{black}{$\frac{\ln\left(8\beta V/3\right)}{\ln|R(T)-R_n(T)|}\geq 0$} if the condition $8\beta V<3$ holds.
Let simply set $\beta=2$. If we further have that $|R(T)-R_n(T)|\leq V\leq 3/16$, the upper bound in Theorem \ref{mainfour} will be the same as that in Theorem \ref{mainone} except for a faster convergence rate\textcolor{black}{.}
Thus, the upper bound in Theorem \ref{mainfour} can be much tighter than that in Theorem \ref{mainone} in the sense that it converges much faster.
\end{remk}

\begin{remk}
The generalization bound in Theorem \ref{mainone} is of order $\mathcal{O}\left(\left(mk\ln(mkn)/n\right)^{\frac{1}{2}}\right)$; while the generalization bound in Theorem \ref{mainfour} is
of order $\mathcal{O}\left(\left(mk\ln(mkn)/n\right)^{\gamma_n}\right)$, where $\gamma_n>1/2$ when $n$ is finite. The generalization bound in Theorem \ref{mainfour}, derived by employing Bennett's inequality, converges faster when the sample size $n$ is small, which is often the case in practice and more detailedly describes the non-asymptotic behavior of the learning process. More empirical discussions can be found in \citet{Zhanguai}.
However, when the sample size $n$ goes to infinity, the term  \textcolor{black}{$\frac{1}{2-\frac{\ln\left(8\beta V/3\right)}{\ln|R(T)-R_n(T)|}}$} will approach to $\frac{1}{2}$, which means that the upper bounds in Theorems \ref{mainfour} and \ref{mainone} describe the same asymptotic behavior of the learning process.
\end{remk}

\begin{remk}
Theorem \ref{mainfour} looks complex, since the exponent in the convergence rate depents itself on the sample size in an implicit way.
Here we show the superiority of Theorem \ref{mainfour} by comparing it with Theorem \ref{mainone}.
From the proof of Theorem \ref{mainfour}, we can see that the theorem depends on the following inequality (\ref{concen}):
$$P\left\{\left|R(T)-R_n(T)\right|\geq\epsilon\right\}\leq2\exp\left(-nVh\left(\frac{\epsilon}{V}\right)\right)\leq2\exp\left(-\beta n\epsilon^{2-\frac{\ln\left(8\beta V/3\right)}{\ln\epsilon}}\right),$$
where $\epsilon\leq V$.
Note that for Hoeffding's inequality, with any $\beta$ we also have
$$P\left\{\left|R(T)-R_n(T)\right|\geq\epsilon\right\}\leq2\exp\left(-2n\epsilon^2\right)=2\exp\left(-\beta n\epsilon^{2-\frac{\ln(\beta/2)}{\ln\epsilon}}\right).$$
Thus, according to Hoeffding's inequality and the prove method of Theorem \ref{mainfour}, for all $T\in\mathcal{T}$, with probability at least $1-\delta$ it holds that
\begin{eqnarray*}
|R(T)-R_n(T)|\leq\frac{2}{n} +\left(\frac{mk\ln\left(4(r+ck)\sqrt{m}ckn\right)+\ln{\frac{2}{\delta}}}{\beta n}\right)^{\frac{1}{2-\frac{\ln(\beta/2)}{\ln|R(T)-R_n(T)|}}}.
\end{eqnarray*}
Comparing the above bound with that in Theorem \ref{mainfour}, we can see that, if we interpret Theorem \ref{mainone} with a faster convergence rate, the upper bound therein is looser than that in Theorem \ref{mainfour} when $V\leq 3/16$.
\end{remk}

Our main results in Theorems \ref{mainone}, \ref{mainthree}, and \ref{mainfour} apply to all the $k$-dimensional coding schemes because the covering number in Lemma 1 measures the complexity of the loss function class that includes all the possible loss functions of $k$-dimensional coding schemes. However, for some specific $k$-dimensional coding schemes, the complexity of the corresponding induced loss function class can be refined. We discuss the details in the next section\footnote{Even though the faster convergence interpretation in Theorem \ref{mainfour} is interesting, it looks complicated and the upper bound is almost the same tight as that of Theorem \ref{mainthree}. Therefore, we do not disscuss its applicaitons for specific $k$-dimensional codeing schemes.}.

\section{Applications}\label{section3}
In this section, we apply our proof methods to specific $k$-dimensional coding schemes. We show that our methods provide state-of-the-art dimensionality-dependent generalization bounds.
\subsection{Non-negative matrix factorization}
NMF factorizes a data matrix $X\in\mathbb{R}_+^{m\times n}$ into two non-negative matrices $T\in\mathbb{R}_+^{m\times k}$ and $Y\in\mathbb{R}_+^{k\times n}$, where $k<\min(m,n)$. NMF has been widely exploited since \citet{lee99} provided a powerful psychological and physiological interpretation as a parts-based factorization and an efficient multiplicative update rule for obtaining a local solution. Many fast and robust algorithms are then followed \citep[see, e.g.,][]{gillis2013fast}. In all applications, both the data points and the vectors $Te_i,i=1,\ldots,k$ are contained in the positive orthant of a finite-dimensional space. In this case, our method for deriving dimensionality-dependent generalization bounds is likely to be superior to the method for obtaining dimensionality-independent results.

Letting $X=(x_1,\ldots,x_n)\in\mathbb{R}_+^{m\times n}$, NMF can be formulated as follows:
\begin{eqnarray}\label{nmfproblem}
&\min_{T, Y}&\|X-TY\|_F^2,\nonumber\\
&\text{s.t.}&T\in\mathbb{R}_+^{m\times k}, Y\in\mathbb{R}_+^{k\times n}\nonumber
\end{eqnarray}
where $\|\cdot\|_F$ is the matrix Frobenius norm.

Because $TY=TQ^{-1}QY$ if $Q$ is a scaling matrix, we can normalize $T$ without changing the optimization problem by choosing
\begin{equation*}
Q=\left(
  \begin{array}{cccc}
    \|T_1\| &  &  &  \\
     & \|T_2\| &  &  \\
     &  & \ddots &  \\
     &  &  & \|T_k\| \\
  \end{array}
\right).
\end{equation*}

If we restrict $\mu\in\mathcal{P}(r)$ and normalize $T$, columns of $Y$ will also be upper bounded by $r$. This can be seen in the following lemma, which generalizes Lemma 2 in \citet{KMaurerP10}:
\begin{lema}\label{l2}
For NMF with normalized $T$, if $\mu\in\mathcal{P}(r)$, then every column of $Y$ is upper bounded by $r$; that is $\|y\|\leq r$ for all $y\in Y$.
\end{lema}


For a fixed $T$, $Y$ is determined by a convex problem. Thus, the reconstruction error for NMF is
\[f_T(x)=\min_{y\in\mathbb{R}_+^{k}}\|x-Ty\|^2,\]
and the generalization error of NMF can be analyzed under the framework of the $k$-dimensional coding schemes.

Using the same proof method as that of Lemma \ref{l1}, we have the following lemma.
\begin{lema}\label{l3}
Let $\mu\in\mathcal{P}(1)$ and $F_\mathcal{T}=\{f_T|T\in\mathcal{T}, \mathcal{T}=\mathbb{R}_+^{m\times k}\}$ be the loss function class induced by the reconstruction error of NMF. We have
\[\ln\mathcal{N}_1(F_\mathcal{T},\xi',n)\leq mk\ln\left(\frac{2(1+k)\sqrt{m}k}{\xi'}\right).\]
\end{lema}

Then, according to the proof methods of Theorems \ref{mainone}, \ref{mainthree} and \ref{mainfour}, we have the following dimensionality-dependent generalization bounds for NMF.

\begin{thm}\label{5}
For NMF, assume that $\mu\in\mathcal{P}(1)$ and that $\mathcal{T}$ is normalized. For any $\delta\in(0,1)$, with probability at least $1-\delta$ it holds for all $T\in\mathcal{T}$ that
\begin{align*}
&|R(T)-R_n(T)| \\
&\leq\frac{2}{n}+\min\left\{\sqrt{\frac{mk\ln\left(2(1+k)\sqrt{m}kn\right)+\ln{2/\delta}}{2n}},\right.\\
&\frac{5\left(mk\ln\left(2(1+k)\sqrt{m}kn\right)+\ln{2/\delta}\right)}{n}\left.+\sqrt{\frac{2R_n(T)\left(mk\ln\left(2(1+k)\sqrt{m}kn\right)+\ln{2/\delta}\right)}{n}}\right\}.
\end{align*}
\end{thm}

\begin{figure}
    \centering
    \begin{subfigure}[b]{0.48\textwidth}
        \includegraphics[width=1\textwidth]{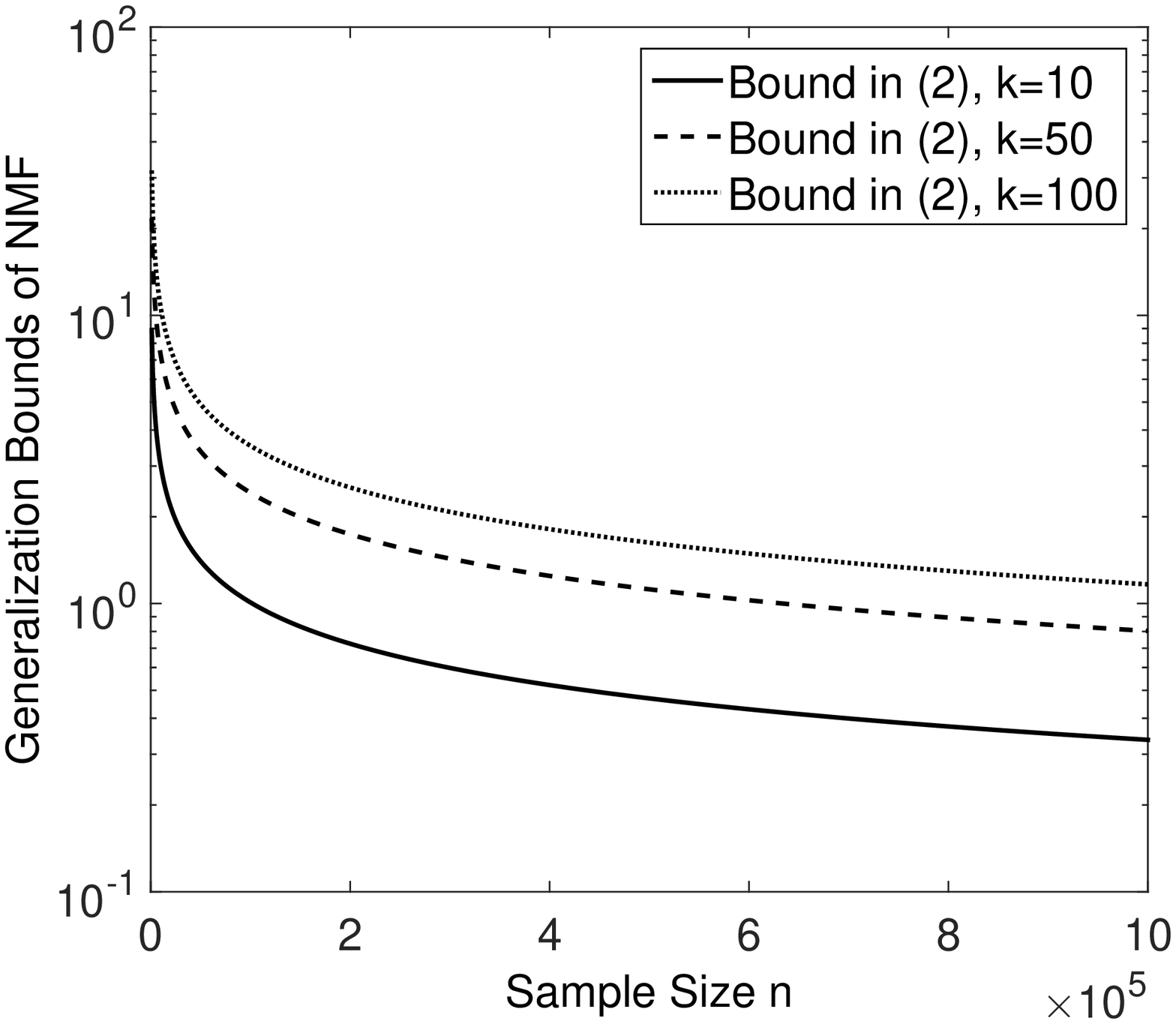}
        \caption{}
        \label{fig:gull}
    \end{subfigure}
    ~ 
    \begin{subfigure}[b]{0.48\textwidth}
        \includegraphics[width=1\textwidth]{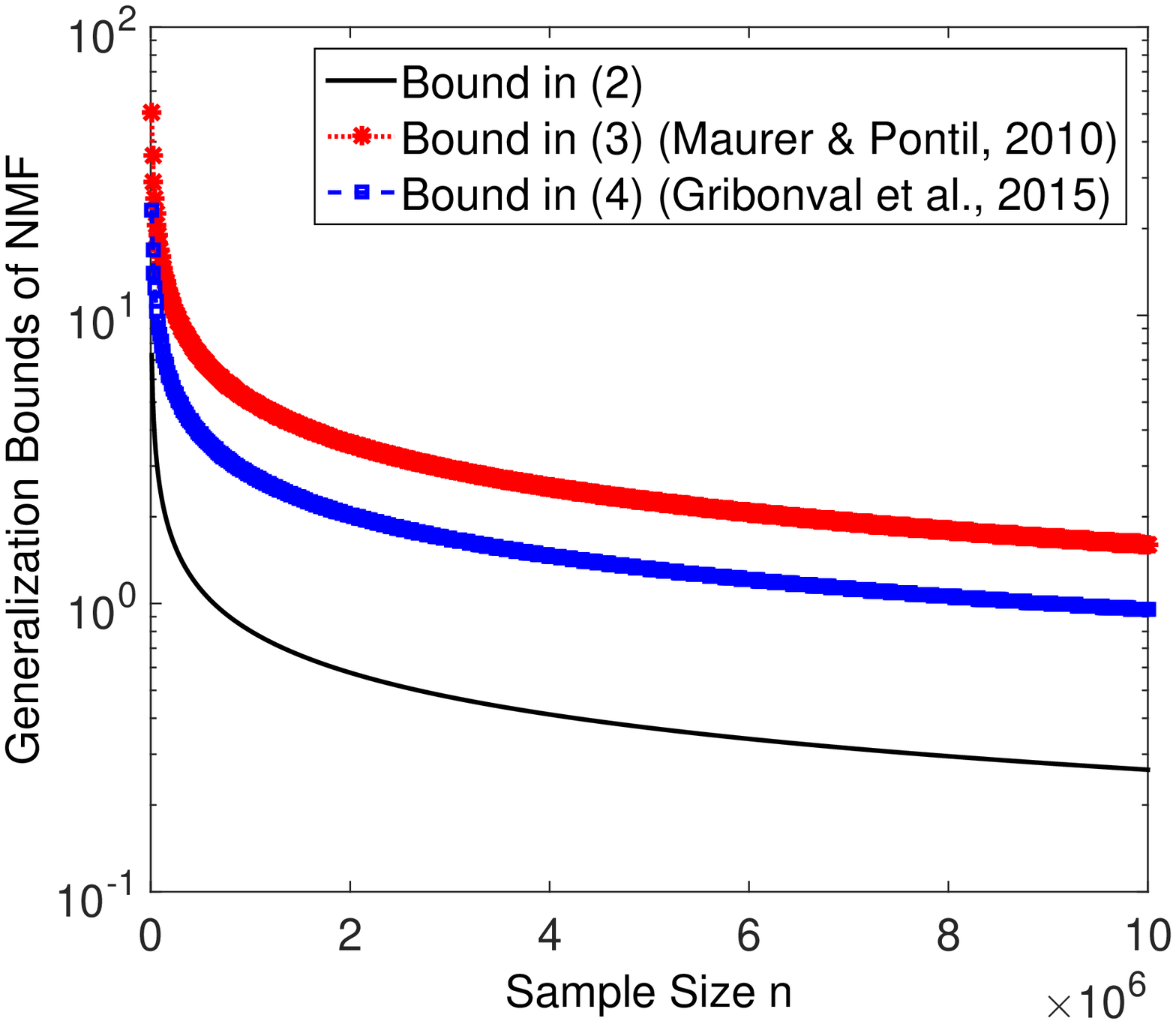}
        \caption{}
        \label{fig:tiger}
    \end{subfigure}
    ~ 
    \begin{subfigure}[b]{0.48\textwidth}
        \includegraphics[width=1\textwidth]{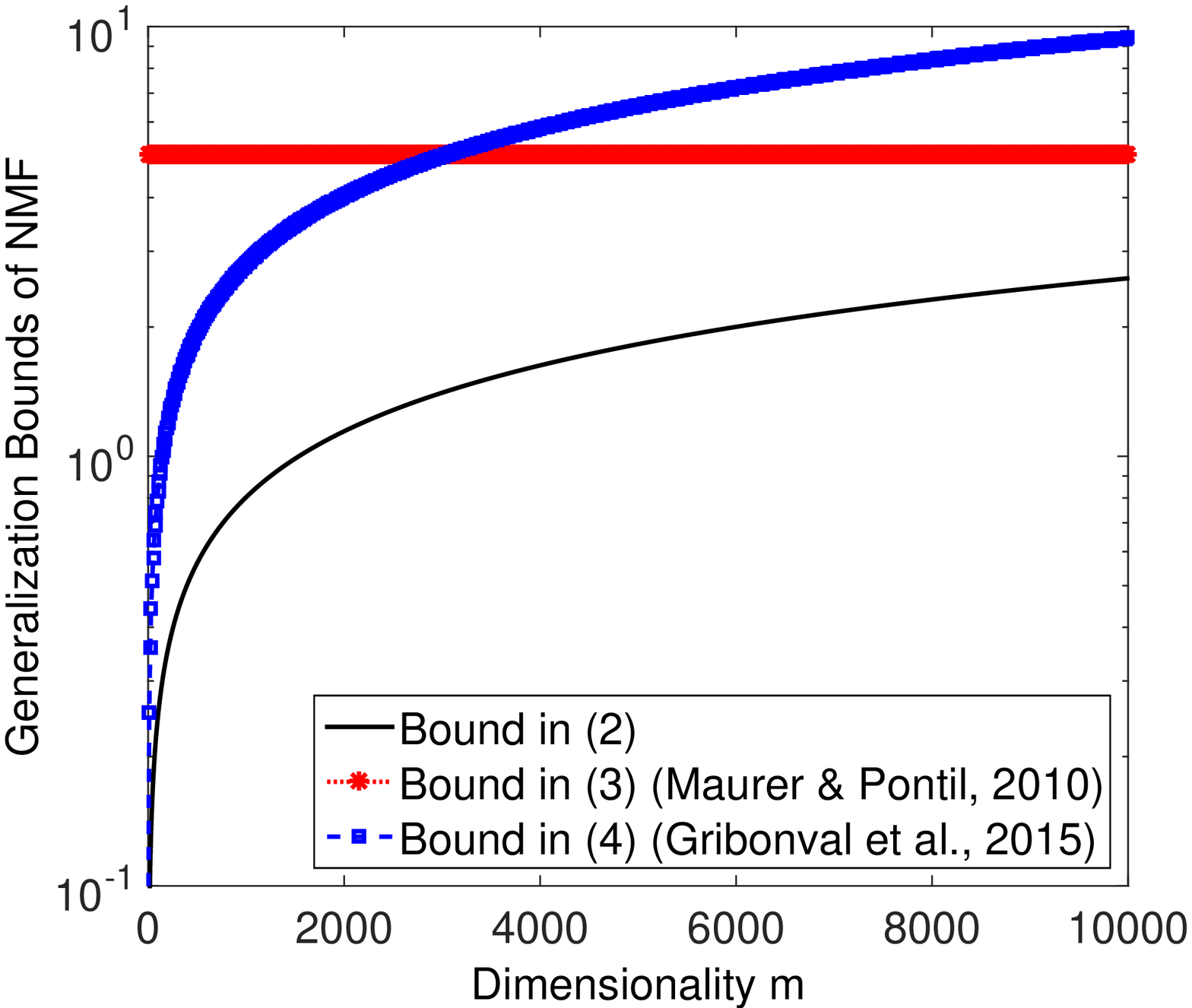}
        \caption{}
        \label{fig:mouse}
    \end{subfigure}
     ~ 
    \begin{subfigure}[b]{0.48\textwidth}
        \includegraphics[width=1\textwidth]{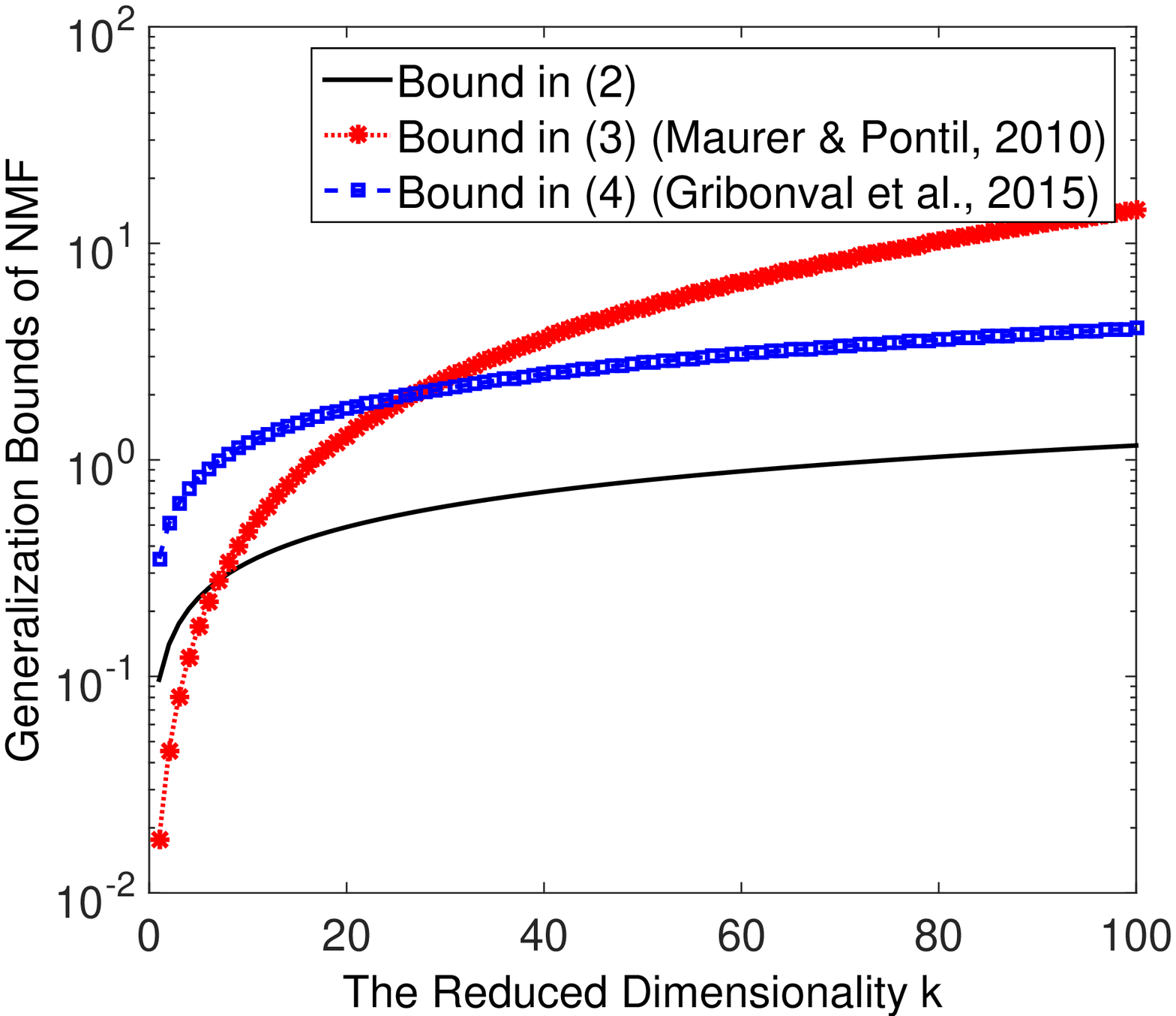}
        \caption{}
        \label{f14}
    \end{subfigure}
    \caption{Comparisons of the generalization bounds of NMF. (a) The convergence of the bound in (\ref{nmfours}), where $m=1000$. (b) Comparing the convergence with state-of-the-art generalization bounds, where $k=50, m=1000$. (c) Comparing the generalization bound with state-of-the-art generalization bounds in terms of the parameter $m$, where $k=50, n=10^6$. (d) Comparing the generalization bound with state-of-the-art generalization bounds in terms of the parameter $k$, where $m=10^3,n=10^6$.}\label{f1}
\end{figure}

Since the value of $R_n(T)$ is unknown in this paper (it is usually known in an optimization procedure), in the rest of the paper, we will only compare the bound in Theorem \ref{mainone} with state-of-the-art bounds. Theorem \ref{mainone} gives the following bound for NMF
\begin{eqnarray}\label{nmfours}
\frac{2}{n}+\sqrt{\frac{mk\ln\left(2(1+k)\sqrt{m}kn\right)+\ln{2/\delta}}{2n}}.
\end{eqnarray}
Under the setting of Theorem \ref{5}, Theorem \ref{pontiltwo} yields the following bound
\begin{eqnarray}\label{nmfpontil}
\frac{k}{\sqrt{n}}\left(14\sqrt{k}+\frac{1}{2}\sqrt{\ln(16nk)}\right)+\sqrt{\frac{\ln 2/\delta}{2n}};
\end{eqnarray}
\citet{gribonval2013sample}'s result gives the following bound
\begin{eqnarray}\label{nmfgib}
\frac{3}{\sqrt{8}}\sqrt{\frac{mk \ln(12\sqrt{8mk})\ln n}{n}}+\frac{1}{\sqrt{8}}\sqrt{\frac{mk\ln(12\sqrt{8mk})+\ln2/\delta}{n}}.
\end{eqnarray}

We then carefully compare the above generalization bounds. For NMF problems, the dimensionality $m$ is usually very large compared to the reduced dimensionality $k$. We set $m=1000, k=50, \delta=0.01$. The comparisons are illustrated in Figure \ref{f1}. The figure shows that in most cases, the derived generalization bound is tighter than state-of-the-art bounds. In Figure \ref{f14},
the bound in (3) is tighter than the derived bound in a small range because it is dimensionality-independent and $m=1000$ is set to be much larger than the corresponding reduced dimensionality $k$.

\subsection{Dictionary learning}
Dictionary learning tries to find a dictionary such that all observed data points can be approximated by linear combinations of atoms in the dictionary. Let the columns of $T$ be the atoms of the dictionary; for an observation $x\in\mathbb{R}^m$, the dictionary learning method will represent $x$ by a linear combination of columns of $T$ as
\[x'= \sum\limits_{i=1}^{k}\alpha_iT_i, \alpha_i\in\mathbb{R},i=1,\ldots,k.\]
Thus, the reconstruction error of dictionary learning is the same as those of $k$-dimensional coding schemes.

\citet{Vainsencher} provided notable dimensionality-dependent generalization bounds for dictionary learning by considering two types of constraints on coefficient selection, respectively. For the $\ell_0$-norm regularized coefficient selection, where every signal is approximated by a combination of, at most, $p$ dictionary atoms, the generalization bound (Theorem 14 therein) is of order $\mathcal{O}(\sqrt{mk\ln(np)/n})$ under an approximate orthogonality assumption on the dictionary. For the $\ell_1$-norm regularized coefficient selection, the generalization bound (Theorem 7 therein) is of order $\mathcal{O}(\sqrt{mk\ln(n\lambda)/n})$ under the requirements that $\lambda$, which is the upper bound of the $\ell_1$-norm of the coefficient, is larger than $e/4$, and that the signal $x$ is mapped onto the $(m-1)$-sphere. Our result on $k$-dimensional coding scheme can also be applied to dictionary learning and provides a more general bound, which does not require $x$ to be on the $(m-1)$-sphere or the near-orthogonality requirement and directly applies to all dictionary learning problems.
\begin{thm}\label{6}
For dictionary learning, assume that $\mu\in\mathcal{P}(1)$ and that $Y$ is a closed subset of the unit ball of $\mathbb{R}^k$, and that every atom $T_i,i=1,\ldots,k$ is bounded by $\|T_i\|\leq c,i=1,\ldots,k$. Then, for any $\delta\in(0,1)$, with probability at least $1-\delta$ it holds for all $T\in\mathcal{T}$ that
\begin{align*}
&|R(T)-R_n(T)| \\
&\leq\frac{2}{n}+\min\left\{\sqrt{\frac{mk\ln\left(4(1+ck)\sqrt{m}ckn\right)+\ln{2/\delta}}{2n}},\right.\\
&\frac{5\left(mk\ln\left(4(1+ck)\sqrt{m}ckn\right)+\ln{2/\delta}\right)}{n}\left.+\sqrt{\frac{2R_n(T)\left(mk\ln\left(4(1+ck)\sqrt{m}ckn\right)+\ln{2/\delta}\right)}{n}}\right\}.
\end{align*}
\end{thm}

The proof of Theorem \ref{6} is the same as that of Theorem \ref{5}.

\begin{remk}
If we substitute an upper bound $\lambda\leq \sqrt{k}$ into the bound in \citet{Vainsencher}, the bound in Theorem 7 therein will be of order $\mathcal{O}(\sqrt{mk\ln(kn)/n})$, which has the same order as term $\sqrt{\frac{mk\ln\left(4(1+ck)\sqrt{m}ckn\right)+\ln{2/\delta}}{2n}}$. However, \textcolor{black}{our bound in Theorem \ref{mainfour} also shows a faster convergence rate.}
\end{remk}

\begin{remk}
The method \citet{Vainsencher} used to upper bound the covering number of the induced loss function class is very different from ours. To upper bound the covering number of the induced loss function class for dictionary learning, \citet{Vainsencher} used the knowledge that a uniform $L$ Lipschitz mapping between metric spaces converts $\xi/L$ covers into $\xi$ covers. Then, they focused on analyzing the Lipschitz property of the reconstruction error function that maps a dictionary into a reconstruction error, i.e, $\Psi_\lambda:D\mapsto h_{R_\lambda,D}, R_\lambda=\{a:\|a\|_1\leq\lambda\}$, as shown in Lemma 7 therein. Also note that to upper bound the Lipschitz constant of the mapping $\Phi_k:D\mapsto h_{H_k,D},H_k=\{a:\|a\|_0\leq k\}$, they introduced the approximate orthogonality condition (a bound on the Babel function) on the dictionary.
\end{remk}

\begin{remk}
Analyzing the Lipschitz properties of the induced loss functions is essential for upper bounding the generalization error of $k$-dimensional coding schemes. Different form the method used in \citet{Vainsencher}, \citet{KMaurerP10} employed Slepian's Lemma to exploit the Lipschitz property; while in this paper, we also proposed a novel method as presented in the proof of Theorem \ref{mainone}.
\end{remk}

The comparisons of the generalization bounds of dictionary learning are similar to that of NMF because NMF can be regarded as dictionary learning in the positive orthant. We therefore omit the comparison. Many algorithms used in applications require sparsity in $Y$, because sparsity has advantages, such as for computation and storage. We therefore analyze sparsity in the next subsection.

\subsection{Sparse coding}
Sparse coding requires sparsity in the codebook. We use the hard constraint discussed in \citet{KMaurerP10}, that is $\mathcal{T}=\{T:\mathbb{R}^k\rightarrow \mathbb{R}^m|\|Te_i\|\leq c,i=1,\ldots,k\}$, $Y=\{y|y\in\mathbb{R}^k, \|y\|_p\leq s\}$, and $1/p+1/q=1, 2\leq p \leq\infty$. Thus, we have
\begin{eqnarray*}
&\|Ty\|&=\left\|\sum\limits_{i=1}^{k}y_iTe_i\right\|\leq \sum\limits_{i=1}^{k}|y_i|\|Te_i\|\\
&&\ \ \ \text{(Using H\"{o}lder's inequality)}\\
&&\leq s\left(\sum\limits_{i=1}^{k}\|Te_i\|^q\right)^{1/q}\leq sck^{1/q}=sck^{1-1/p}.
\end{eqnarray*}

The following generalization bound for sparse coding is also from the work of \citet{KMaurerP10}, derived using the proof method of Theorem \ref{pontiltwo}.
\begin{thm}\label{7}
For sparse coding, assume that $\mu\in\mathcal{P}(1)$. Let $Y=\{y|y\in\mathbb{R}^k, \|y\|_p\leq s\}$ where $1\leq p \leq\infty$. Let also assume that for all $T\in\mathcal{T}$, $\|Te_i\|\leq 1,i=1,\ldots,k$. Then, for any $\delta\in(0,1)$, with probability at least $1-\delta$ it holds for all $T\in\mathcal{T}$ that
\begin{eqnarray*}
\left|R(T)-R_n(T)\right|\leq  \frac{k}{2}\sqrt{\frac{\ln{(16ns^22k^{2-2/p})}}{n}}+\sqrt{\frac{\ln{2/\delta}}{2n}}+\frac{4+4sk^{1-1/p}+\sqrt{8\pi}sk^{2-1/p}}{\sqrt{n}}.
\end{eqnarray*}
\end{thm}

We now consider the generalization bound of sparse coding using our method. The following lemma is proved in Section \ref{proofl4}.
\begin{lema}\label{boundcoversparse}
Follow the setting of Theorem \ref{7}. Let $F_\mathcal{T}$ be the loss function class of sparse coding. We have
\begin{eqnarray*}
&&\ln\mathcal{N}_1(F_\mathcal{T},\xi',n)\leq mk\ln\left(\frac{4(s+s^2k^{1-1/p})\sqrt{m}k^{1-1/p}}{\xi'}\right).
\end{eqnarray*}
\end{lema}

Then, we have the generalization bounds for sparse coding as follows:
\begin{thm}\label{sparsecoding}
Follow the setting of Theorem \ref{7}. For any $\delta\in(0,1)$, with probability at least $1-\delta$ it holds for all $T\in\mathcal{T}$ that
\begin{eqnarray*}
&&|R(T)-R_n(T)|\\
&& \leq\min\left\{\frac{2}{n}+\sqrt{\frac{\Delta+\ln{2/\delta}}{2n}},\right.\left.\frac{2}{n}+\frac{5\left(\Delta+\ln{2/\delta}\right)}{n}+\sqrt{\frac{2R_n(T)\left(\Delta+\ln{2/\delta}\right)}{n}}\right\},
\end{eqnarray*}
where $\Delta=mk\ln\left(4(s+s^2k^{1-1/p})\sqrt{m}k^{1-1/p}n\right)$.
\end{thm}

The proof of Theorem \ref{sparsecoding} is the same as that of Theorem \ref{5}.

Theorem \ref{sparsecoding} gives the following bound for sparse coding
\begin{eqnarray}\label{sparse1}
\frac{2}{n}+\sqrt{\frac{mk\ln\left(4(s+s^2k^{1-1/p})\sqrt{m}k^{1-1/p}n\right)+\ln{2/\delta}}{2n}}.
\end{eqnarray}
The upper bound for sparse coding derived by \citet{KMaurerP10} is presented in Theorem \ref{7}:
\begin{eqnarray}\label{sparse2}
 \frac{k}{2}\sqrt{\frac{\ln{(16ns^22k^{2-2/p})}}{n}}+\sqrt{\frac{\ln{2/\delta}}{2n}}+\frac{4+4sk^{1-1/p}+\sqrt{8\pi}sk^{2-1/p}}{\sqrt{n}}.
\end{eqnarray}
\citet{gribonval2013sample}'s result gives the following bound for sparse coding.
\begin{eqnarray}\label{sparse3}
\frac{1}{\sqrt{8}}\left(3\sqrt{\frac{mk \max\left(\ln\left(6\sqrt{8}sk^{1-1/p}\right),1\right)\ln n}{n}}+\sqrt{\frac{mk\max\left(\ln\left(6\sqrt{8}sk^{1-1/p}\right),1\right)+\ln2/\delta}{n}}\right).
\end{eqnarray}

\begin{figure}
    \centering
    \begin{subfigure}[b]{0.48\textwidth}
        \includegraphics[width=1\textwidth]{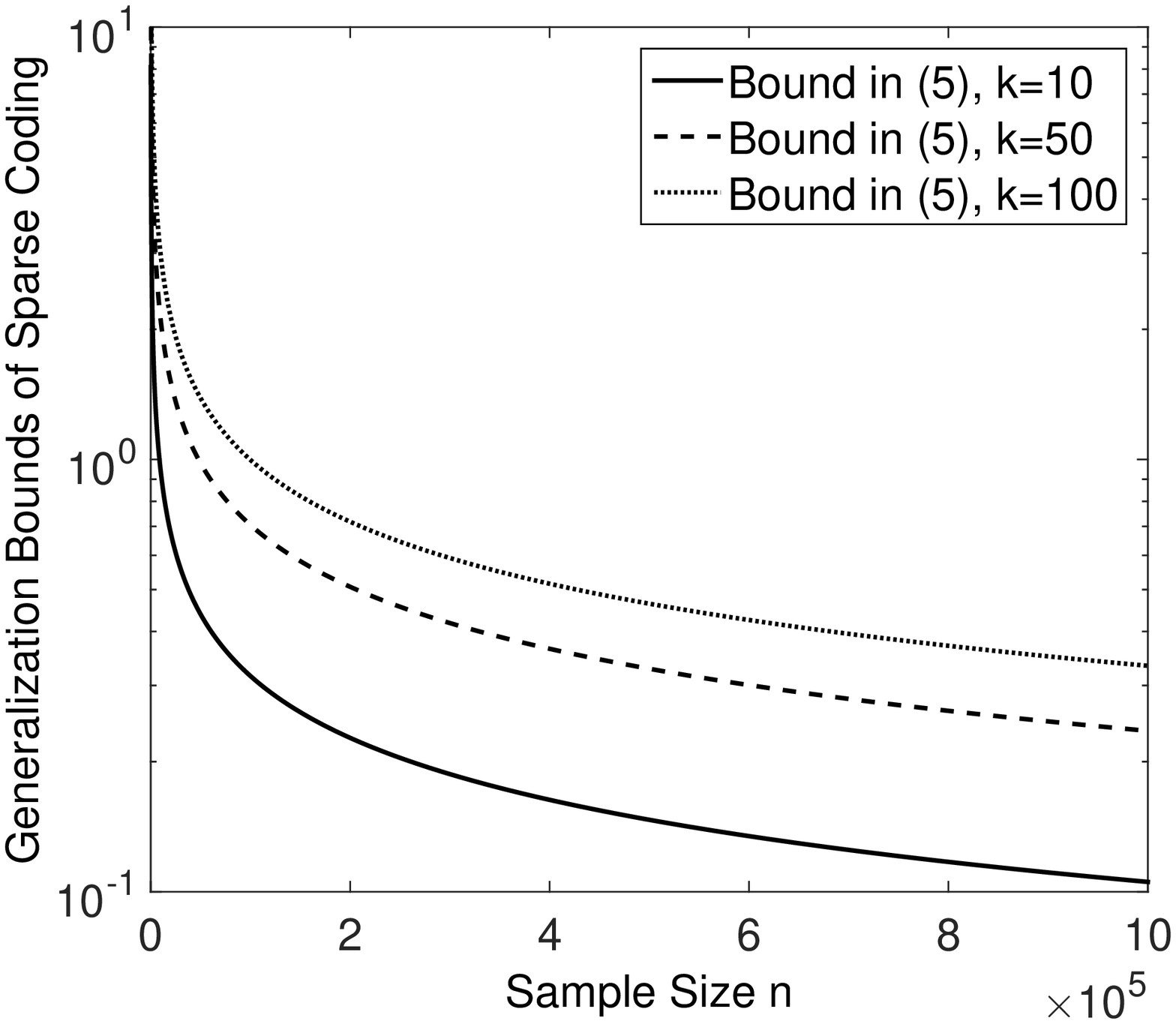}
        \caption{}
        \label{fig:gull}
    \end{subfigure}
    ~ 
    \begin{subfigure}[b]{0.48\textwidth}
        \includegraphics[width=1\textwidth]{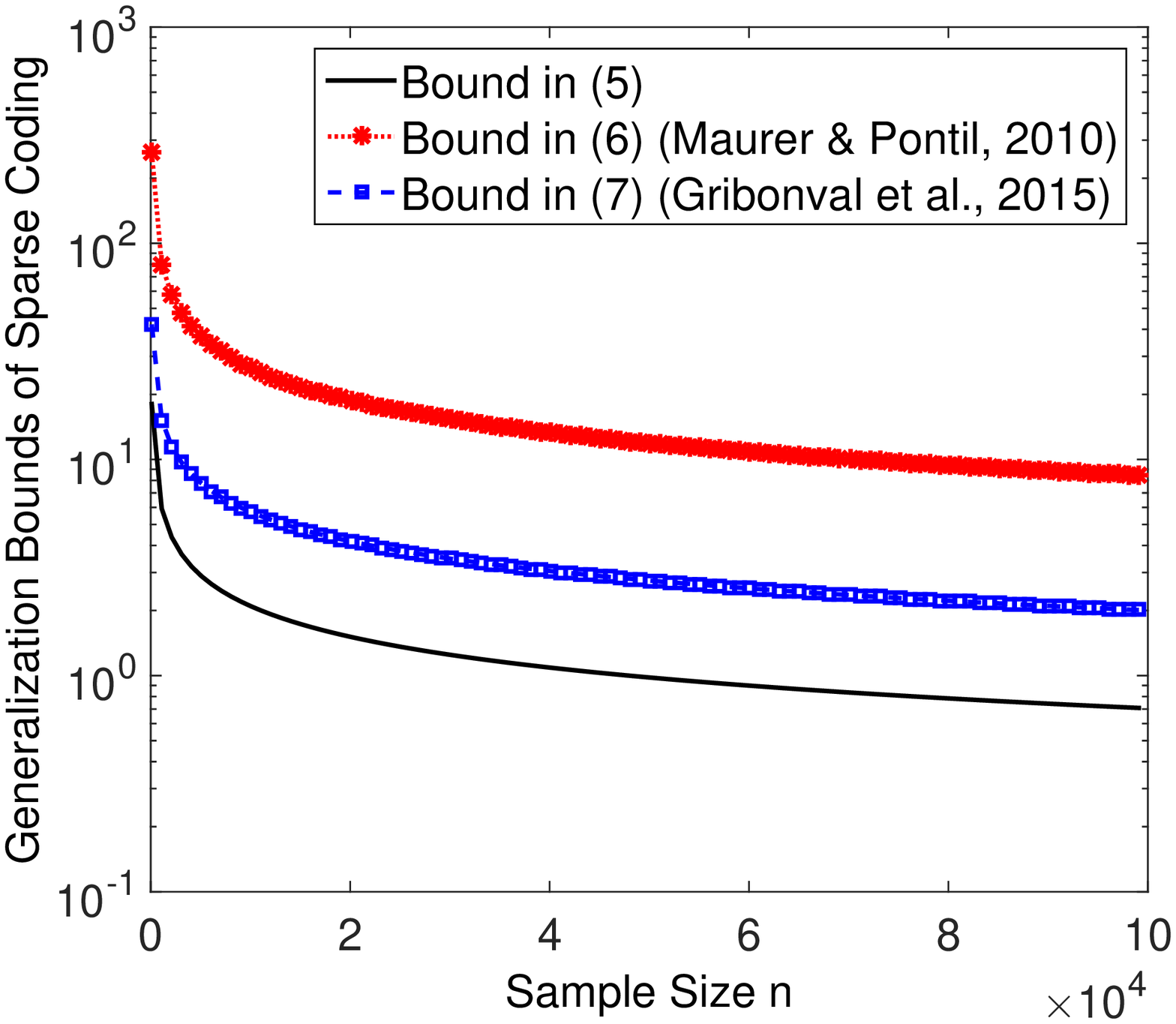}
        \caption{}
        \label{fig:tiger}
    \end{subfigure}
    ~ 
    \begin{subfigure}[b]{0.48\textwidth}
        \includegraphics[width=1\textwidth]{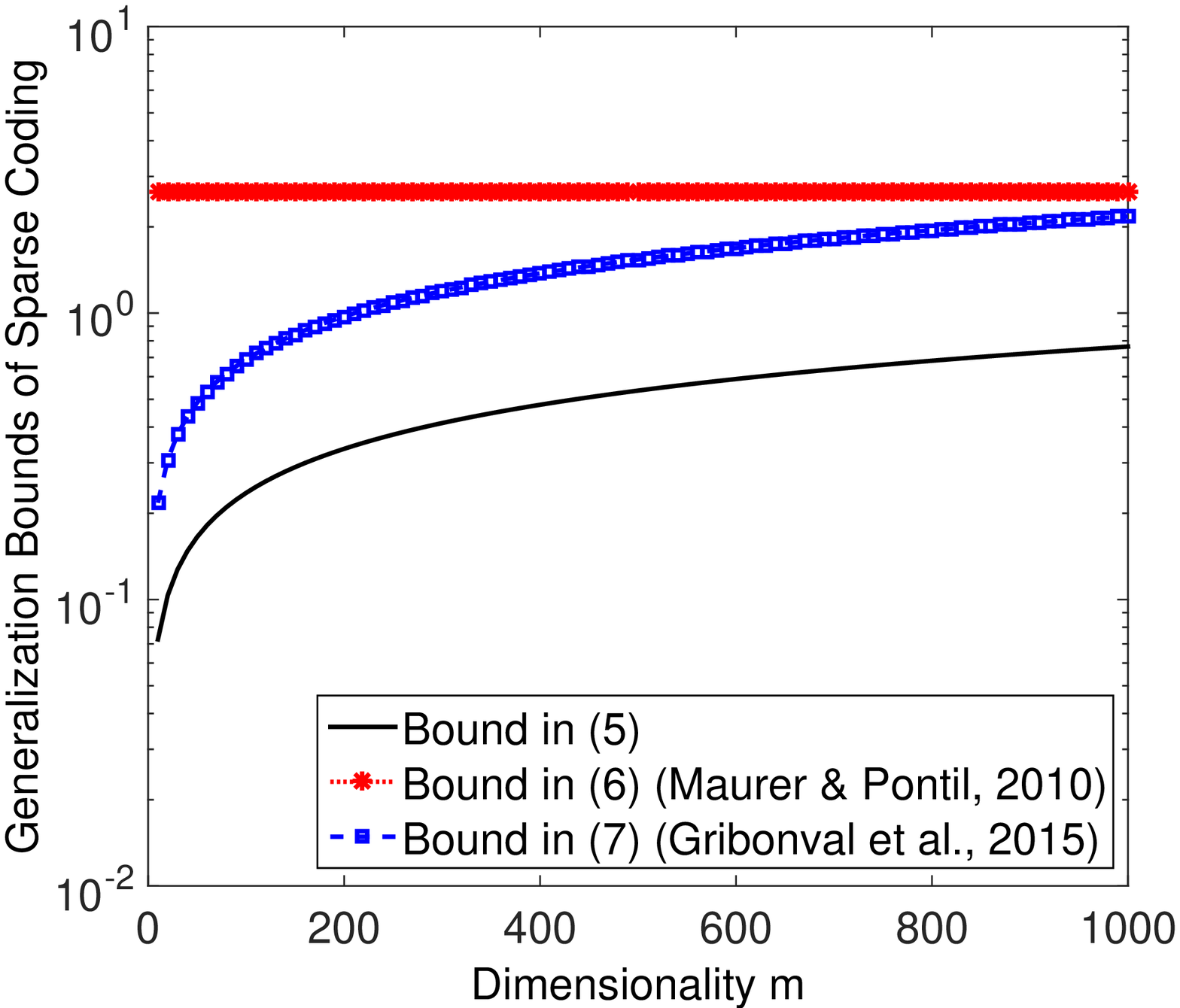}
        \caption{}
        \label{fig:mouse}
    \end{subfigure}
     ~ 
    \begin{subfigure}[b]{0.48\textwidth}
        \includegraphics[width=1\textwidth]{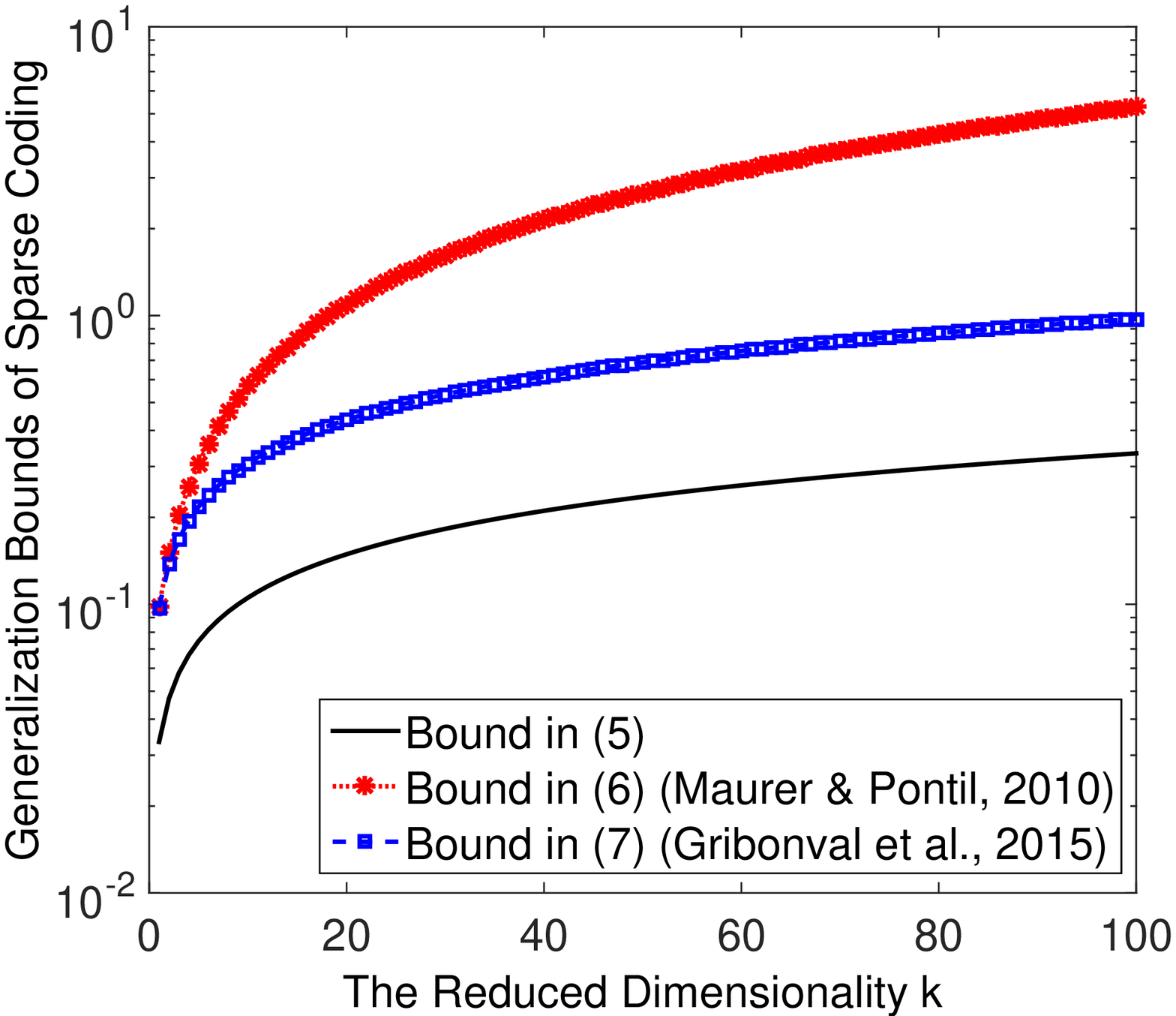}
        \caption{}
        \label{fig:mouse}
    \end{subfigure}
    \caption{Comparisons of the generalization bounds of sparse coding. (a) The convergence of the bound in (\ref{sparse1}), where $m=100$. (b) Comparing the convergence with state-of-the-art generalization bounds, where $k=50, m=100$. (c) Comparing the generalization bound with state-of-the-art bounds in terms of the parameter $m$, where $k=50, n=10^6$. (d) Comparing the generalization bound with state-of-the-art bounds in terms of the parameter $k$, where $m=100, n=10^6$.}\label{f2}
\end{figure}

We then compare the above generalization bounds of sparse coding in Figure \ref{f2} by setting $m=100, k=50, \delta=0.01, p=1$, and $s=10$. The comparisons show that the derived generalization bound is tighter than state-of-the-art bounds.

\subsection{Vector quantization and $k$-means clustering}
The $k$-means clustering (or vector quantization) method aims to find $k$ cluster centers such that observations can be partitioned into $k$ clusters and represented by the $k$ cluster centers with a small reconstruction error. Taking every column of $T$ as a cluster center and setting $Y$ as \textcolor{black}{the standard bases} $\{e_1,\ldots,e_k\}$, we see that solving a $k$-means clustering problem is equal to finding an implementation $T$. The corresponding reconstruction error is
\[f_T(x)=\min_{i\in\{1,\ldots,k\}}\|x-Te_i\|^2.\]
So, the reconstruction error of $k$-means clustering and vector quantization is also within the framework of the reconstruction error of $k$-dimensional coding schemes.




The following lemma is essential for proving our dimensionality-dependent generalization bounds.
\begin{lema}\label{l5}
Assume that $\mu\in\mathcal{P}(1)$. Let $F_\mathcal{T}$ be the loss function class of $k$-means clustering and vector quantization. Then
\[\ln\mathcal{N}_1(F_\mathcal{T},\xi',n)\leq mk\ln\left(\frac{8\sqrt{m}}{\xi'}\right).\]
\end{lema}

\begin{thm}\label{k-means}
For $k$-means clustering and vector quantization, assume that $\mu\in\mathcal{P}(1)$, and that the functions $f_T$ for $T\in\mathcal{T}$ have \textcolor{black}{a} range contained in $[0,1]$. Then, for any $\delta\in(0,1)$, with probability at least $1-\delta$ it holds for all $T\in\mathcal{T}$ that
\begin{eqnarray*}
&&|R(T)-R_n(T)|\leq\frac{2}{n}+\min\left\{\sqrt{\frac{mk\ln\left(8\sqrt{m}n\right)+\ln{2/\delta}}{2n}},\right.\\
&&\ \ \ \ \ \ \frac{5\left(mk\ln\left(8\sqrt{m}n\right)+\ln{2/\delta}\right)}{n}\left.+\sqrt{\frac{2R_n(T)\left(mk\ln\left(8\sqrt{m}n\right)+\ln{2/\delta}\right)}{n}}\right\}.
\end{eqnarray*}
\end{thm}

The proof of Theorem \ref{k-means} is the same as that of Theorem \ref{5}.

Theorem \ref{k-means} gives the following bound for $k$-means clustering and vector quantization
\begin{eqnarray}\label{kmeans1}
\frac{2}{n}+\sqrt{\frac{mk\ln\left(8\sqrt{m}n\right)+\ln{2/\delta}}{2n}}.
\end{eqnarray}
\citet{KMaurerP10} derived the following bound
\begin{eqnarray}\label{kmeans2}
\frac{3\sqrt{2\pi}k r^2}{\sqrt{n}}+r^2\sqrt{\frac{8\ln{1/\delta}}{n}}.
\end{eqnarray}

\citet{gribonval2013sample} provided the following bound
\begin{eqnarray}\label{kmeans3}
\frac{3}{\sqrt{8}}\sqrt{\frac{mk \ln(12\sqrt{8})\ln n}{n}}+\frac{1}{\sqrt{8}}\sqrt{\frac{mk\ln(12\sqrt{8})+\ln2/\delta}{n}}.
\end{eqnarray}

\begin{remk}\label{remark13}
The bound in (\ref{kmeans2}) has order $\mathcal{O}(k/\sqrt{n})$, which is the same as the bound obtained by \citet{biau}.
The term $\sqrt{\frac{mk\ln\left(8\sqrt{m}nr^2\right)+\ln{2/\delta}}{2n}}$ in Theorem \ref{k-means} has order $\mathcal{O}(\sqrt{mk\ln{(mn)}/n})$. If $m\ln{(mn)}\leq k$, our bound can be tighter than that of \citet{KMaurerP10} and the result in \citet{biau}. The generalization bounds derived by \citet{KMaurerP10} and \citet{biau} also have an advantage that they converge faster. As discussed in \citet{Bartminmax}, \citet{Linder}, and \citet{Devroye}, the factor $\sqrt{\ln{n}}$ in Theorem \ref{k-means} can be removed by the sophisticated uniform large-deviation inequalities of \citet{Alexander} or \citet{Talagrand94}. However, \citet{Devroye} proved that (Theorem 12.10 therein) the fast convergence upper bound has an astronomically large constant. The corresponding convergence bound is therefore loose. Our generalization bound, which is derived by exploiting Bennett's inequality, will be tighter if the empricial reconstruction error $R_n(T)$ is small.
\end{remk}

\begin{figure}
    \centering
    \begin{subfigure}[b]{0.48\textwidth}
        \includegraphics[width=1\textwidth]{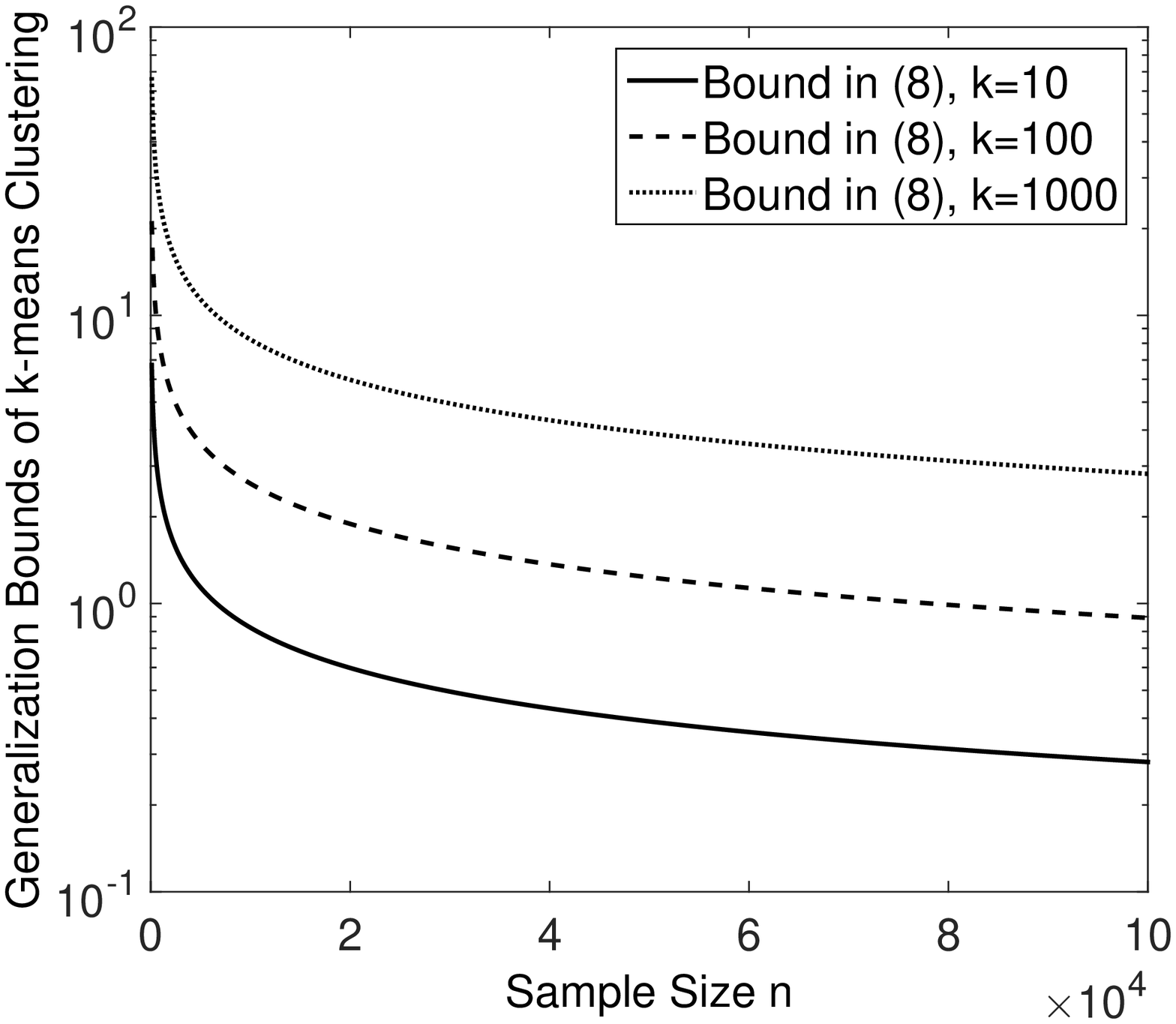}
        \caption{}
        \label{fig:gull}
    \end{subfigure}
    ~ 
    \begin{subfigure}[b]{0.48\textwidth}
        \includegraphics[width=1\textwidth]{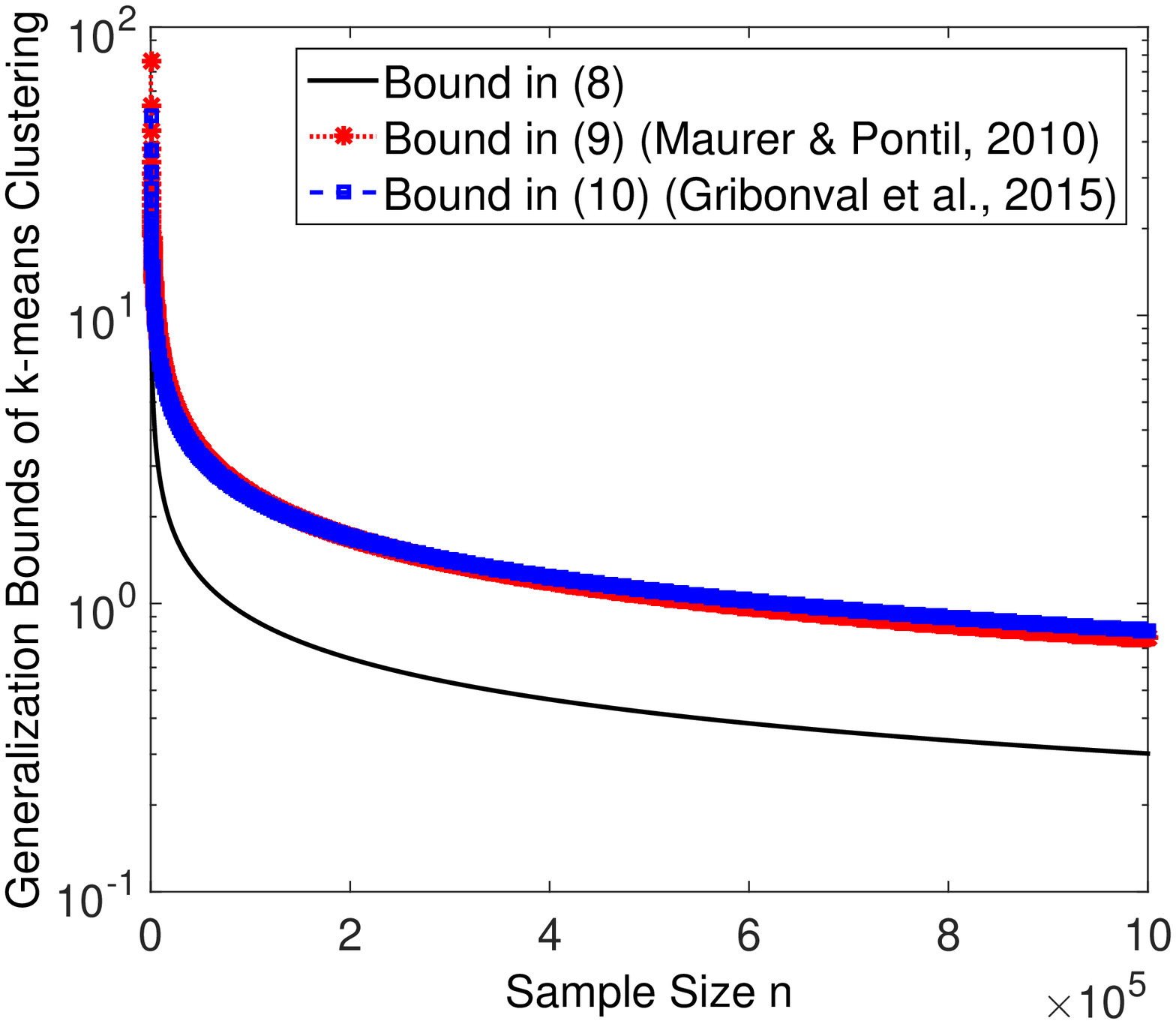}
        \caption{}
        \label{fig:tiger}
    \end{subfigure}
    ~ 
    \begin{subfigure}[b]{0.48\textwidth}
        \includegraphics[width=1\textwidth]{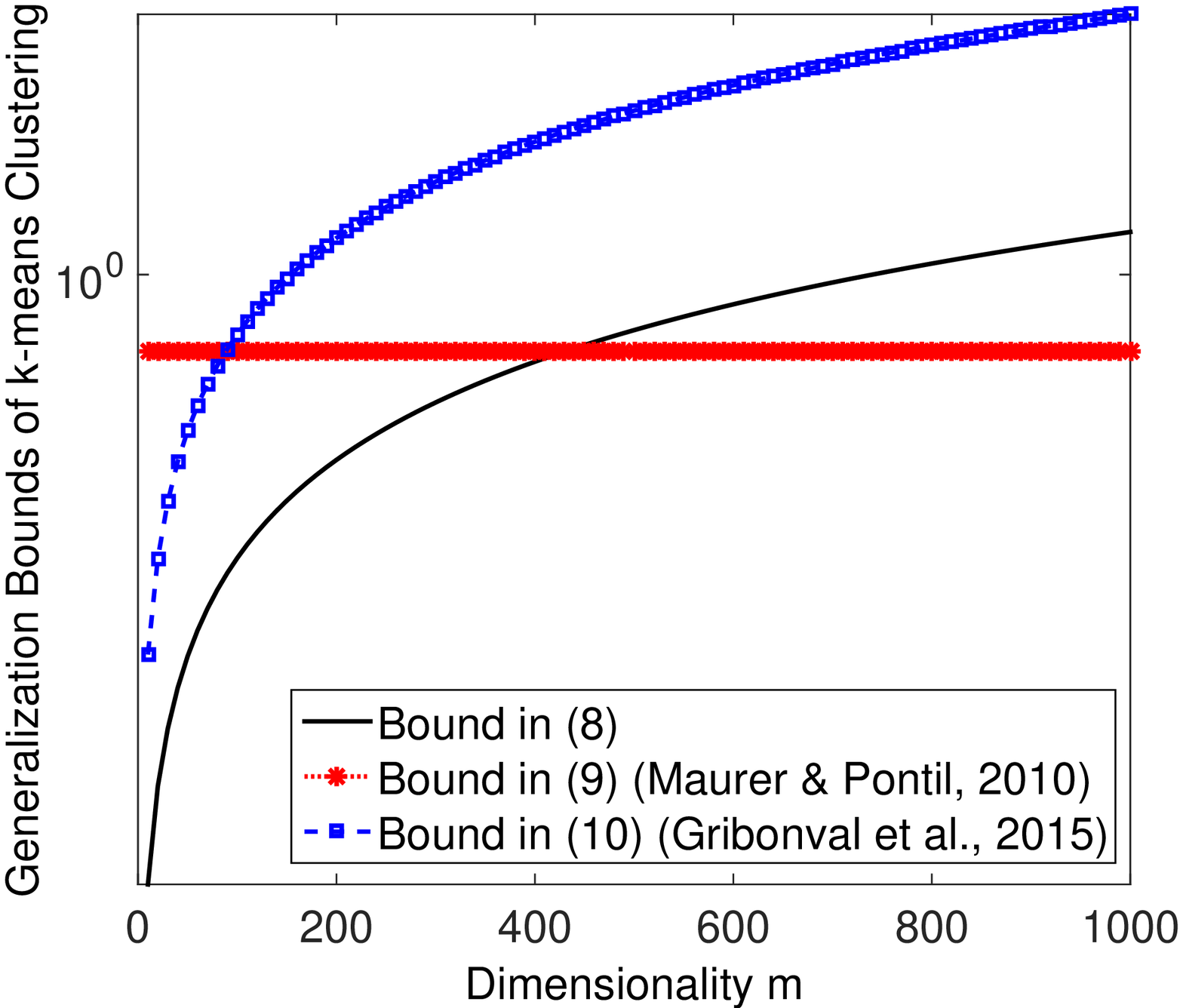}
        \caption{}
        \label{fig:mouse}
    \end{subfigure}
     ~ 
    \begin{subfigure}[b]{0.48\textwidth}
        \includegraphics[width=1\textwidth]{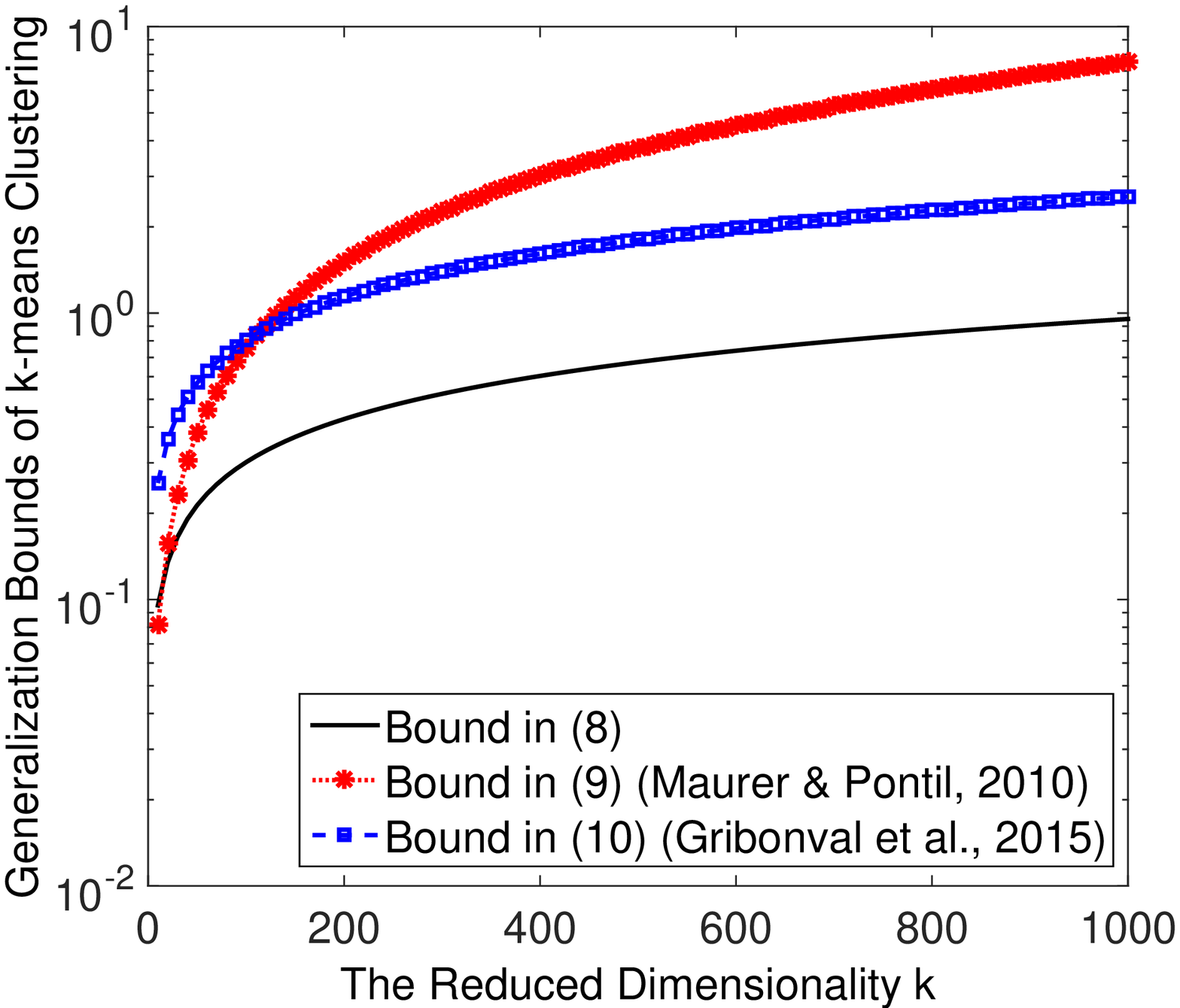}
        \caption{}
        \label{fig:mouse}
    \end{subfigure}
    \caption{Comparisons of the generalization bounds of $k$-means clustering and vector quantization. (a) The convergence of the bound in (\ref{kmeans1}), where $m=100$. (b) Comparing the convergence with state-of-the-art generalization bounds, where $k=m=100$. (c) Comparing the generalization bound with state-of-the-art bounds in terms of the parameter $m$, where $k=100, n=10^{6}$. (d) Comparing the generalization bound with state-of-the-art bounds in terms of the parameter $k$, where $m=100, n=10^6$.}\label{f3}
\end{figure}

We compare the above generalization bounds of $k$-means clustering and vector quantization in Figure \ref{f3} by setting $k=m=100$. For $k$-means clustering and vector quantization problems, the dimensionality $m$ can be independent of the reduced dimensionality $k$. Figure \ref{f3} shows that when $k$ is not very large, the derived bound is tighter than state-of-the-art generalization bounds.

\section{Proofs}\label{section5}

In this section we prove the main results in Section 2 and some of the results presented in Section 3.
\subsection{Concentration inequalities}
In this subsection, we introduce the concentration inequalities that will be used to prove our assertions.

We first present Hoeffding's inequality \citep{hoeffding1963probability}, which is widely used for deriving generalization bounds.
\begin{thm}[Hoeffding's inequality]\label{hoeffding}
Let $X=\{x_1,\ldots,x_n\}$ $\in\mathcal{H}^n$ be a sample set of independent random variables such that $x_i\leq B$ for some $B> 0$ almost surely for all $i\leq n$. Then for any $X\in\mathcal{H}^n$ and $\epsilon>0$, the following inequality holds:
\begin{eqnarray*}
P\left\{\left|E\frac{1}{n}\sum_{i=1}^{n}x_i-\frac{1}{n}\sum_{i=1}^{n}x_i\right|\geq\epsilon\right\}\leq2\exp\left(\frac{-2n\epsilon}{B^2}\right).
\end{eqnarray*}
\end{thm}


We will also use Bernstein's inequality and Bennett's inequality \citep{boucheron2013concentration,Zhanguai} to derive generalization bounds.
\begin{thm}[Bernstein's inequality]\label{bernstein}
Let $X=\{x_1,\ldots,x_n\}$ $\in\mathcal{H}^n$ be a sample set of independent random variables such that $x_i\leq B$ for some $B> 0$ and $Ex_i^2$ is no bigger than $V$ for some $V>0$ almost surely for all $i\leq n$. Then for any $X\in\mathcal{H}^n$ and $\epsilon>0$, the following inequality holds:
\begin{eqnarray*}
&&P\left\{\left|E\frac{1}{n}\sum_{i=1}^{n}x_i-\frac{1}{n}\sum_{i=1}^{n}x_i\right|\geq\epsilon\right\}\leq2\exp\left(\frac{-n\epsilon^2}{2(V+B\epsilon/3)}\right).
\end{eqnarray*}
\end{thm}

\begin{thm}[Bennett's inequality]\label{bennett}
Let $X=\{x_1,\ldots,x_n\}$ $\in\mathcal{H}^n$ be a sample set of independent random variables such that $x_i\leq B$ for some $B> 0$ and $Ex_i^2$ is no bigger than $V$ for some $V>0$ almost surely for all $i\leq n$. Then for any $X\in\mathcal{H}^n$ and $\epsilon>0$, the following inequality holds:
\begin{eqnarray*}
&&P\left\{\left|E\frac{1}{n}\sum_{i=1}^{n}x_i-\frac{1}{n}\sum_{i=1}^{n}x_i\right|\geq\epsilon\right\}\leq2\exp\left(-\frac{nV}{B^2}h\left(\frac{B\epsilon}{V}\right)\right),
\end{eqnarray*}
where $h(x)=(1+x)\ln(1+x)-x$ for $x>0$.
\end{thm}

\subsection{Proof of Lemma \ref{l1}}
\emph{Proof.}
We will bound the covering number of the loss function class $F_\mathcal{T}$ by bounding the covering number of the implementation class $\mathcal{T}$. Cutting the subspace $[-c,c]^m\subset\mathbb{R}^m$ into small $m$-dimensional regular solids with width $\xi$, there are a total of
\[\left\lceil\frac{2c}{\xi}\right\rceil^m\leq \left(\frac{2c}{\xi}+1\right)^m\leq \left(\frac{4c}{\xi}\right)^m\]
such regular solids. If we pick out the centers of these regular solids and use them to make up $T$, there are
\[\left\lceil\frac{2c}{\xi}\right\rceil^{mk}\leq \left(\frac{4c}{\xi}\right)^{mk}\]
choices, denoted by $\mathcal{S}$. Then $|\mathcal{S}|$ is the upper bound of the $\xi$-cover of the implementation class $\mathcal{T}$.

We will prove that for every $T$, there exists a $T'\in \mathcal{S}$ such that
$$\sup_{x}|f_T(x)-f_{T'}(x)|\leq \xi',$$
where $\xi'=(r+ck)\sqrt{m}k\xi$. The proof is as follows:
\begin{eqnarray*}
&&|f_T(x)-f_{T'}(x)|\\
&&=\left|\min_y\|x-Ty\|^2-\min_y\|x-T'y\|^2\right|\\
&&=\left|\min_y\|x-Ty\|^2+\max_y\left(-\|x-T'y\|^2\right)\right|\\
&&\leq\left|\max_y\left(\|x-Ty\|^2-\|x-T'y\|^2\right)\right|\\
&&\leq\left|\max_y2x^\top Ty-2x^\top T'y\right|+\left|\max_y\|Ty\|^2-\|T'y\|^2\right|\\
&&= \left|\max_y\sum\limits_{i=1}^{k}y_i\left<2x,(T-T')e_i\right>\right|+\left|\max_y\sum\limits_{i,j}^{k}y_iy_j\left<(T+T')e_i,(T-T')e_j\right>\right|\\
&&\ \ \ \text{(Using H\"{o}lder's inequality)}\\
&&\leq \left|\sum\limits_{i=1}^{k}|\left<2x,(T-T')e_i\right>|\right| +\left|\sum\limits_{i,j}^{k}|\left<(T+T')e_i,(T-T')e_j\right>|\right|\\
&&\ \ \ \text{(Using Cauchy-Schwarz inequality)}\\
&&\leq \left|\sum\limits_{i=1}^{k}\|2x\|\|(T-T')e_i\|\right| +\left|\sum\limits_{i,j}^{k}\|(T+T')e_i\|\left\|(T-T')e_j\right\|\right|\\
&&\leq \left|\sum\limits_{i=1}^{k}\|2x\|\left\|\frac{\xi}{2}\textbf{1}\right\|\right|+\left|\sum\limits_{i,j}^{k}\|(T+T')e_i\|\left\|\frac{\xi}{2}\textbf{1}\right\|\right|\\
&&\leq\sqrt{m}rk\xi+\sqrt{m}ck^2\xi \\
&&= (r+ck)\sqrt{m}k\xi=\xi'.
\end{eqnarray*}
The last inequality holds because of the triangle inequality. We have
\begin{eqnarray*}
&&\sum\limits_{i,j}^{k}\|(T+T')e_i\|\leq\sum\limits_{i,j}^{k}\left(\|Te_i\|+\|T'e_i\|\right)\leq\sum\limits_{i,j}^{k}2c=2ck^2.
\end{eqnarray*}

Let $F_\mathcal{T}$ denote the loss function class for the algorithms when searching for implementations $T\in\mathcal{T}$ and the metric $d$ be the metric that $d(f_T(x),f_{T'}(x))=\sup_x|f_T(x)-f_{T'}(x)|$. According to Definition \ref{coveringnumber}, for $\forall f_T\in F_\mathcal{T}$, there is a $T'\in \mathcal{S}$ such that
\begin{eqnarray*}
&& \|d(f_T(X),f_{T'}(X))\|_1=\left[\sum\limits_{i=1}^{n}d(f_T(x_i),f_{T'}(x_i))\right]\leq n\xi'.
\end{eqnarray*}
Thus,
\begin{eqnarray*}
&&\mathcal{N}_1(F_\mathcal{T},\xi',n)\leq|\mathcal{S}|\leq \left(\frac{4c}{\xi}\right)^{mk}=\left(\frac{4(r+ck)\sqrt{m}ck}{\xi'}\right)^{mk}.
\end{eqnarray*}
Taking log on both sides, we have
\[\ln\mathcal{N}_1(F_\mathcal{T},\xi',n)\leq mk\ln\left(\frac{4(r+ck)\sqrt{m}ck}{\xi'}\right).\] \hfill$\blacksquare$

\subsection{Proof of Theorem \ref{mainone}}
We first prove the following theorem, which is useful to prove Theorem \ref{mainone}.
\begin{thm}\label{boundhoe}
Let $X=\{x_1,\ldots,x_n\}\sim\mu^n$ be a set of independent random variables such that $f_T(x_i)\leq b$ for some $b> 0$ almost surely for all $f_T\in F_\mathcal{T}$ and $i\leq n$. Then for any $X\sim\mu^n$ and $\delta\in(0,1)$, with probability at least $1-\delta$, we have
\begin{eqnarray*}
&&\sup_{f_T\in F_\mathcal{T}}\left|R(T)-R_n(T)\right|\leq\frac{2}{n}+b\sqrt{\frac{\ln{\mathcal{N}_1(F_\mathcal{T},1/n,n)+\ln{2/\delta}}}{2n}},
\end{eqnarray*}
where $R_n(T)=\frac{1}{n}\sum_{i=1}^{n}f_T(x_i)$ and $R(T)=E_xR_n(T)$.
\end{thm}

\emph{Proof.}
Since $F_T(X)=\{f_T(x_1),\ldots,f_T(x_n)\}$ is a set of independent random variables, according to Hoeffding's inequality, for any $\delta\in(0,1)$, with probability at least $1-\delta$, we have
\begin{eqnarray*}
&&\left|R(T)-R_n(T)\right|\leq b\sqrt{\frac{\ln{2/\delta}}{2n}}.
\end{eqnarray*}
Let $F_{\mathcal{T},\epsilon}$ be a minimal $\epsilon$-cover of $F_T$. Then, $|F_{\mathcal{T},\epsilon}|=\mathcal{N}_1(F_\mathcal{T},\epsilon,n)$. By a union bound of probability, we have that with probability at least $1-\delta$, the following holds
\begin{eqnarray}\label{union1}
\sup_{f_T\in F_{\mathcal{T},\epsilon}}\left|R(T)-R_n(T)\right|\leq b\sqrt{\frac{\ln{2\mathcal{N}_1(F_{\mathcal{T}},\epsilon,n)/\delta}}{2n}}=b\sqrt{\frac{\ln{\mathcal{N}_1(F_{\mathcal{T}},\epsilon,n)}+\ln{2/\delta}}{2n}}.
\end{eqnarray}
It can be easily verified that
\begin{eqnarray}\label{cover2ep}
&&\sup_{f_T\in F_T}|R(T)-R_n(T)|\leq 2\epsilon +\sup_{f_T\in F_{T,\epsilon}}|R(T)-R_n(T)|.
\end{eqnarray}
Combine inequalities (\ref{union1}) and (\ref{cover2ep}), and let $\epsilon=1/n$, we have that with probability at least $1-\delta$, the following holds
\begin{eqnarray*}
&&\sup_{f_T\in F_\mathcal{T}}\left|R(T)-R_n(T)\right|\leq\frac{2}{n}+b\sqrt{\frac{\ln{\mathcal{N}_1(F_\mathcal{T},1/n,n)+\ln{2/\delta}}}{2n}},
\end{eqnarray*}
which concludes the proof. \hfill$\blacksquare$

Theorem \ref{mainone} can be proven by combining Theorem \ref{boundhoe} and Lemma \ref{l1}. We can also prove Proposition 1 using the same method as that of Theorem \ref{mainone}.

\subsection{Proof of Theorem \ref{mainthree}}
According to Bernstein's inequality, we have the following theorem, which is useful to prove Theorem \ref{mainthree}.

\begin{thm}\label{boundbern}
Let $X=\{x_1,\ldots,x_n\}\sim\mu^n$ be a set of independent random variables such that $f_T(x_i)\leq 1$ almost surely for all $f_T\in F_\mathcal{T}$ and $i\leq n$. Then for any $X\sim\mu^n$ and $\delta\in(0,1)$, with probability at least $1-\delta$, we have
\begin{eqnarray*}
&&\sup_{f_T\in F_\mathcal{T}}\left|R(T)-R_n(T)\right|\\
&&\leq\frac{2}{n}+\frac{5\left(\ln\mathcal{N}_1(F_\mathcal{T},1/n,n)+\ln{2/\delta}\right)}{n}+\sqrt{\frac{2R_n(T)\left(\ln{\mathcal{N}_1(F_\mathcal{T},1/n,n)}+\ln{2/\delta}\right)}{n}}.
\end{eqnarray*}
\end{thm}

\emph{Proof.}
Since $F_T(X)=\{f_T(x_1),\ldots,f_T(x_n)\}$ is a set of independent random variables, according to Bernstein's inequality, for any $\delta\in(0,1)$, with probability at least $1-\delta$, we have
\begin{eqnarray}\label{bern11}
&&\left|R(T)-R_n(T)\right|\leq \frac{2\ln{2/\delta}}{3n}+\sqrt{\frac{2V\ln{2/\delta}}{n}}.
\end{eqnarray}
We also have that $V\leq R(T)$ because $Ef_T(x_i)^2\leq Ef_T(x_i)=R(T)$. Collecting the terms in $R(T)$, completing the square and solving for $\sqrt{R(T)}$ shows that with probability at least $1-\delta$, we have
\begin{eqnarray}\label{bern2}
&&\sqrt{R(T)}\leq\sqrt{R_n(T)} +3\sqrt{\frac{\ln{2/\delta}}{n}}.
\end{eqnarray}
Straightforward substitution of inequality (\ref{bern2}) into inequality (\ref{bern11}) shows that with probability at least $1-\delta$, we have
\begin{eqnarray*}
&&\left|R(T)-R_n(T)\right|\leq \frac{5\ln{2/\delta}}{n}+\sqrt{\frac{2R_n(T)\ln{2/\delta}}{n}}.
\end{eqnarray*}
Similar to the proof of Theorem \ref{boundhoe}, by a union bound of probability, we then have that with probability at least $1-\delta$, the following holds
\begin{eqnarray*}
&&\sup_{f_T\in F_\mathcal{T}}\left|R(T)-R_n(T)\right|\\
&&\leq\frac{2}{n}+\frac{5\left(\ln\mathcal{N}_1(F_\mathcal{T},1/n,n)+\ln{2/\delta}\right)}{n} +\sqrt{\frac{2R_n(T)\left(\ln{\mathcal{N}_1(F_\mathcal{T},1/n,n)}+\ln{2/\delta}\right)}{n}}.
\end{eqnarray*}
which concludes the proof. \hfill$\blacksquare$

Theorem \ref{mainthree} can be proven by combining Theorem \ref{boundbern} and Lemma \ref{l1}.
\subsection{Proof of Theorem \ref{mainfour}}
The following theorem, derived by exploiting Bennett's inequality, is essential to prove Theorem \ref{mainfour}.

\begin{thm}\label{boundbenn}
Let $X=\{x_1,\ldots,x_n\}\sim\mu^n$ be a set of independent random variables such that $f_T(x_i)\leq 1$ almost surely for all $f_T\in F_\mathcal{T}$ and $i\leq n$. Then for any $X\sim\mu^n$ and $\delta\in(0,1)$, with probability at least $1-\delta$ it holds for all $T\in\mathcal{T}$ that
\begin{eqnarray*}
&\left|R(T)-R_n(T)\right|&\leq\frac{2}{n}+\left(\frac{\ln{\mathcal{N}_1(F_\mathcal{T},1/n,n)}+\ln{2/\delta}}{\beta n}\right)^{\frac{1}{2-\frac{\ln\left(8\beta V/3\right)}{\ln\left|R(T)-R_n(T)\right|}}}
\end{eqnarray*}
when $V$ is no smaller than $\left|R(T)-R_n(T)\right|$ and there is a positive constant $\beta$ such that $8\beta V<3$.
\end{thm}

Theorem \ref{boundbenn} can be easily proven by using Berenstain's inequality. However, to show the faster convergence propery, we propose a new method to prove Berenstain's inequlity, which needs the following lemma.
\begin{lema}\label{lbenn}
For $\epsilon\in(0,1]$ and $V\geq\epsilon$, there exists some $\beta>0$ and $0<\gamma<2$ 
such that the following holds
\begin{eqnarray*}
-Vnh\left(\frac{\epsilon}{V}\right)\leq -\beta n\epsilon^\gamma\leq O\left(-n\epsilon^2\right).
\end{eqnarray*}
Let $\{x_1,\ldots,x_n\}$ be i.i.d. variables such that $x_i\leq 1$, $Ex_i^2\leq V$ and $\left|R(T)-R_n(T)\right|\leq V$ are almost surely for all $i\leq n$.
Then, for any $\delta\in(0,1)$, with probability at least $1-\delta$, we have
\begin{eqnarray*}
&&\left|R(T)-R_n(T)\right|\leq \left(\frac{\ln2/\delta}{\beta n}\right)^{\frac{1}{2-\frac{\ln\left(8\beta V/3\right)}{\ln\left|R(T)-R_n(T)\right|}}}.
\end{eqnarray*}
\end{lema}

\emph{Proof.}
We prove the first part. We have
\begin{eqnarray*}
&&-Vnh\left(\frac{\epsilon}{V}\right)\leq -\beta n\epsilon^\gamma\\
&\Longleftrightarrow& V\left(\left(1+\frac{\epsilon}{V}\right)\ln\left(1+\frac{\epsilon}{V}\right)-\frac{\epsilon}{V}\right)\geq \beta \epsilon^\gamma\\
&&\ \ \ (\text{Because that $\epsilon<1$})\\
&\Longleftrightarrow& \gamma\geq \frac{\ln\left(\frac{V}{\beta }\left(\left(1+\frac{\epsilon}{V}\right)\ln\left(1+\frac{\epsilon}{V}\right)-\frac{\epsilon}{V}\right)\right)}{\ln{\epsilon}}.
\end{eqnarray*}

It holds that
\begin{eqnarray*}
&&\frac{\ln\left(\frac{V}{\beta }\left(\left(1+\frac{\epsilon}{V}\right)\ln\left(1+\frac{\epsilon}{V}\right)-\frac{\epsilon}{V}\right)\right)}{\ln{\epsilon}}\\
&&\ \ \ \left(\text{Because $(1+x)\ln(1+x)\geq \frac{1}{2+\frac{2x}{3}}x^2+x$ for $x\geq 0$}\right)\\
&&\leq \frac{\ln\left(\frac{V}{\beta }\frac{3}{6+\frac{2\epsilon}{V}}\left(\frac{\epsilon}{V}\right)^2\right)}{\ln{\epsilon}}=\frac{\ln\left(\frac{3\epsilon^2}{\beta(6V+2\epsilon)}\right)}{\ln\epsilon}\\
&&=2-\frac{\ln\left(2\beta (V+\frac{\epsilon}{3})\right)}{\ln\epsilon}\\
&&\leq 2,\ \text{when $\epsilon\leq V$ and $8\beta V<3$}.
\end{eqnarray*}

Thus, there are many pairs of $(\beta,\gamma)$ such that the first part of Lemma \ref{lbenn} holds.

We then prove Berenstain's inequality and the second part. According to Bennett's inequality, we have
\begin{eqnarray}\label{concen}
&\textcolor{black}{P}\left\{\left|R(T)-R_n(T)\right|\geq\epsilon\right\}&\leq2\exp\left(-nVh\left(\frac{\epsilon}{V}\right)\right)\nonumber\\
&&\leq2\exp\left(-\beta n\epsilon^{2-\frac{\ln\left(2\beta (V+\frac{\epsilon}{3})\right)}{\ln\epsilon}}\right)\\
&&=2\exp\left(\frac{-n\epsilon^2}{2(V+\frac{\epsilon}{3})}\right),\nonumber
\end{eqnarray}
which is the Berenstain's inequality.

To prove the second part, let $\epsilon<V$. We have
\begin{eqnarray*}
&\textcolor{black}{P}\left\{\left|R(T)-R_n(T)\right|\geq\epsilon\right\}&\leq2\exp\left(\frac{-n\epsilon^2}{2(V+\frac{\epsilon}{3})}\right)\\
&&\leq2\exp\left(\frac{-n\epsilon^2}{2(V+\frac{V}{3})}\right)\\
&&=2\exp\left(-\beta n\epsilon^{2-\frac{\ln\left(\frac{8\beta V}{3}\right)}{\ln\epsilon}}\right).
\end{eqnarray*}
For any $\delta\in(0,1)$, let
\begin{eqnarray}\label{43}
2\exp\left(-\beta n\epsilon^{2-\frac{\ln\left(\frac{8\beta V}{3}\right)}{\ln\epsilon}}\right)=\delta.
\end{eqnarray}
Then, with probability at least $1-\delta$, we have
\begin{eqnarray}\label{42}
\left|R(T)-R_n(T)\right|\leq \epsilon.
\end{eqnarray}
Combining (\ref{43}) and (\ref{42}), with probability at least $1-\delta$, we have
\begin{eqnarray*}
\frac{\ln2/\delta}{\beta n}=\epsilon^{2-\frac{\ln\left(\frac{8\beta V}{3}\right)}{\ln\epsilon}}\geq\epsilon^{2-\frac{\ln\left(\frac{8\beta V}{3}\right)}{\ln\left|R(T)-R_n(T)\right|}}
\end{eqnarray*}
and
\begin{eqnarray}\label{44}
\epsilon\leq \left(\frac{\ln2/\delta}{\beta n}\right)^{\frac{1}{2-\frac{\ln\left(\frac{8\beta V}{3}\right)}{\ln\left|R(T)-R_n(T)\right|}}}.
\end{eqnarray}
Combining (\ref{42}) and (\ref{44}), with probability at least $1-\delta$, we have
\begin{eqnarray*}
\left|R(T)-R_n(T)\right|\leq \left(\frac{\ln2/\delta}{\beta n}\right)^{\frac{1}{2-\frac{\ln\left(\frac{8\beta V}{3}\right)}{\ln\left|R(T)-R_n(T)\right|}}}.
\end{eqnarray*}
Thus, the Second part of Lemma \ref{lbenn} holds. \hfill$\blacksquare$

Similar to the proof of Theorem \ref{boundhoe}, Theorem \ref{boundbenn} can be proven by using Lemma \ref{lbenn} and a union bound of probability.

Theorem \ref{mainfour} can be proven by combining Theorem \ref{boundbenn} and Lemma \ref{l1}.

\subsection{Proof of Lemma \ref{l2}}
The proof method is the same as that of Lemma 2 in \citep{KMaurerP10}.

\emph{Proof.} Let
\[h(y)=\left\|x-\sum\limits_{i=1}^{k}T_iy_i\right\|^2.\]
Assume that $y$ is a minimizer of $h$ and $\|y\|>r$. Because $T$ is normalized, $\|T_i\|=1,i,\ldots,k$. Then
\[\left\|\sum\limits_{i=1}^{k}T_iy_i\right\|^2=\|y\|^2+\sum_{i\neq j}y_iy_j\left<T_i,T_j\right>>r^2.\]

\noindent Let the real-valued function $f$ be defined as
\[f(t)=h(ty).\]
Then
\begin{eqnarray*}
&&f'(1)=2\left(\left\|\sum\limits_{i=1}^{k}T_iy_i\right\|^2-\left<x,\sum\limits_{i=1}^{k}T_iy_i\right>\right)\\
&&\ \ \ \text{(Using Cauchy-Schwarz inequality)}\\
&&\geq 2\left(\left\|\sum\limits_{i=1}^{k}T_iy_i\right\|^2-r\left\|\sum\limits_{i=1}^{k}T_iy_i\right\|\right)\\
&&=2\left(\left\|\sum\limits_{i=1}^{k}T_iy_i\right\|-r\right)\left\|\sum\limits_{i=1}^{k}T_iy_i\right\|>0.
\end{eqnarray*}
So $f$ cannot have a minimum at $1$, whence $y$ cannot be a minimizer of $h$. Thus, the minimizer $y$ must be contained in the ball with radius $r$ in the $m$-dimensional space. \hfill$\blacksquare$

\subsection{Proof of Lemma \ref{boundcoversparse}}\label{proofl4}

\emph{Proof.}
As in the proof of Lemma \ref{l1}, we can pick out a set $\mathcal{S}$, where $|\mathcal{S}|\leq \left(\frac{4c}{\xi}\right)^{mk}$, having the property that for every $T$, there exists a $T'\in \mathcal{S}$ such that $\sup_x|f_T(x)-f_{T'}(x)|\leq \xi'$ with $\xi'=(rs+cs^2k^{1-1/p})\sqrt{m}\xi k^{1-1/p}$. The detail is as follows.
\begin{eqnarray}\label{part0}
&&|f_T-f_{T'}|=\left|\min_y\|x-Ty\|^2-\min_y\|x-T'y\|^2\right|\nonumber\\
&&\leq\left|\max_y\left(\|x-Ty\|^2-\|x-T'y\|^2\right)\right|\nonumber\\
&&\leq\left|\max_y2x^\top Ty-2x^\top T'y\right|+\left|\max_y\|Ty\|^2-\|T'y\|^2\right|\\
&&=\left|\max_y\sum\limits_{i=1}^{k}y_i\left<2x,(T-T')e_i\right>\right|+\left|\max_y\sum\limits_{i,j}^{k}y_iy_j\left<(T+T')e_i,(T-T')e_j\right>\right|\nonumber.
\end{eqnarray}

Using H\"{o}lder's inequality, we have
\begin{eqnarray}\label{part1}
&&\left|\max_y\sum\limits_{i=1}^{k}y_i\left<2x,(T-T')e_i\right>\right|\nonumber\\
&&\leq \left|\max_y\|y\|_p\left(\sum\limits_{i=1}^{k}\left|\left<2x,(T-T')e_i\right>\right|^q\right)^{1/q}\right|\\
&&\leq \left|\max_y\|y\|_p\left(\sum\limits_{i=1}^{k}\left|\|2x\|\|(T-T')e_i\|\right|^q\right)^{1/q}\right|\nonumber\\
&&\leq \sqrt{m}sr\xi k^{1/q}\nonumber\\
&&\leq \sqrt{m}sr\xi k^{1-1/p}.\nonumber
\end{eqnarray}

Using H\"{o}lder's inequality again, we have inequalities (\ref{big1}) and (\ref{big2}):
\begin{eqnarray}\label{big1}
&&\left|\max_y\sum\limits_{i,j}^{k}y_iy_j\left<(T+T')e_i,(T-T')e_j\right>\right|\nonumber\\
&&\leq\left|\max_y\|y\|_p\left(\sum\limits_{i}^{k}\left|\sum\limits_{j}^{k}\left<(T+T')e_i,(T-T')e_j\right>y_j\right|^q\right)^{1/q}\right|,
\end{eqnarray}
and
\begin{eqnarray}\label{big2}
&&\left|\sum\limits_{j}^{k}\left<(T+T')e_i,(T-T')e_j\right>y_j\right|\nonumber\\
&&\leq\left|\left(\sum\limits_{j}^{k}\left<(T+T')e_i,(T-T')e_j\right>^q\right)^{1/q}\left(\sum\limits_{j}^{k}|y_j|^p\right)^{1/p}\right|\nonumber\\
&&\leq\left|\left(\sum\limits_{j}^{k}\left(\|(T+T')e_i\|\|(T-T')e_j\|\right)^q\right)^{1/q}\left(\sum\limits_{j}^{k}|y_j|^p\right)^{1/p}\right|\nonumber\\
&&\leq\left|\left(\sum\limits_{j}^{k}\left((\|Te_i\|+\|T'e_i\|)\|(T-T')e_j\|\right)^q\right)^{1/q}\left(\sum\limits_{j}^{k}|y_j|^p\right)^{1/p}\right|\\
&&\leq\sqrt{m}sc\xi k^{1/q}=\sqrt{m}sc\xi k^{1-1/p}.\nonumber
\end{eqnarray}

Combining inequalities (\ref{big1}) and (\ref{big2}), it gives
\begin{eqnarray}\label{part2}
&&\left|\max_y\sum\limits_{i,j}^{k}y_iy_j\left<(T+T')e_i,(T-T')e_j\right>\right|\nonumber\\
&&\leq \left|\max_y\|y\|_p\left(\sum\limits_{i}^{k}\left|\sqrt{m}sc\xi k^{1-1/p}\right|^q\right)^{1/q}\right|\\
&&\leq \sqrt{m}s^2c\xi k^{2-2/p}.\nonumber
\end{eqnarray}

Combining inequalities (\ref{part0}), (\ref{part1}) and (\ref{part2}), we have
\begin{eqnarray*}
&&|f_T-f_{T'}|\leq\left|\max_y\sum\limits_{i=1}^{k}y_i\left<2x,(T-T')e_i\right>\right|\\
&&\ \ \ +\left|\max_y\sum\limits_{i,j}^{k}y_iy_j\left<(T+T')e_i,(T-T')e_j\right>\right|\\
&&\leq \sqrt{m}sr\xi k^{1-1/p} +\sqrt{m}s^2c\xi k^{2-2/p}\\
&&=(rs+cs^2k^{1-1/p})\sqrt{m}\xi k^{1-1/p}=\xi'.
\end{eqnarray*}
According to Definition \ref{coveringnumber}, for $\forall f_T\in F_\mathcal{T}$, there is a $T'\in \mathcal{S}$ such that
\begin{eqnarray*}
&&\|d(f_T(X),f_{T'}(X))\|_1=\left[\sum\limits_{i=1}^{2}d(f_T(x_i),f_{T'}(x_i))\right]\leq 2\xi'.
\end{eqnarray*}
Thus,
\begin{eqnarray*}
&&\mathcal{N}_1(F_\mathcal{T},\xi',n)\leq|\mathcal{S}|\leq \left(\frac{4c}{\xi}\right)^{mk}=\left(\frac{4(rs+cs^2k^{1-1/p})\sqrt{m}ck^{1-1/p}}{\xi'}\right)^{mk}.
\end{eqnarray*}
Taking log on both sides, we have
\begin{eqnarray*}
&&\ln\mathcal{N}_1(F_\mathcal{T},\xi',n)\leq mk\ln\left(\frac{4(rs+cs^2k^{1-1/p})\sqrt{m}ck^{1-1/p}}{\xi'}\right),
\end{eqnarray*}
which concludes the proof. \hfill$\blacksquare$

\subsection{Proof of Lemma \ref{l5}}
The proof method of Lemma \ref{l5} is similar to that of Lemma \ref{l1}.

\emph{Proof.}
For $k$-means clustering and vector quantization, we can easily prove that $\|Te_i\|\leq r, i=1,\ldots,k$. As in the proof of Lemma \ref{l1} and Lemma \ref{boundcoversparse}, we can pick out a set $\mathcal{S}$, where $|\mathcal{S}|\leq \left(\frac{4r}{\xi}\right)^{mk}$, having the property that for every $T$ there exists a $T'\in \mathcal{S}$ such that $\sup_x|f_T(x)-f_{T'}(x)|\leq \xi'$ with $\xi'=2r\sqrt{m}\xi$. The proof is as follows:
\begin{eqnarray*}
&&|f_T-f_{T'}|\\
&&\leq\left|\max_{i\in\{1,\ldots,k\}}\left(\|x-Te_i\|^2-\|x-T'e_i\|^2\right)\right|\\
&&\leq\left|\max_{i\in\{1,\ldots,k\}}2x^\top Te_i-2x^\top T'e_i\right| +\left|\max_{i\in\{1,\ldots,k\}}\|Te_i\|^2-\|T'e_i\|^2\right|\\
&&=\left|\max_{i\in\{1,\ldots,k\}}\left<2x,(T-T')e_i\right>\right| +\left|\max_{i\in\{1,\ldots,k\}}\left<(T+T')e_i,(T-T')e_i\right>\right|\\
&&\ \ \ \text{(Using Cauchy-Schwarz inequality)}\\
&&\leq\left|\max_{i\in\{1,\ldots,k\}}\|2x\|\|(T-T')e_i\|\right| +\left|\max_{i\in\{1,\ldots,k\}}\left(\|Te_i\|+\|T'e_i\|\right)\|(T-T')e_i\|\right|\\
&&\leq\sqrt{m}r\xi+\sqrt{m}r\xi\\
&&=2r\sqrt{m}\xi=\xi'.
\end{eqnarray*}

Thus,
\begin{eqnarray*}
\ \ \ \mathcal{N}_1(F_\mathcal{T},\xi',n)\leq|\mathcal{S}|\leq \left(\frac{4r}{\xi}\right)^{mk}=\left(\frac{8r^2\sqrt{m}}{\xi'}\right)^{mk}.
\end{eqnarray*}
Taking log on both sides, we have
\[\ln\mathcal{N}_1(F_\mathcal{T},\xi',n)\leq mk\ln\left(\frac{8r^2\sqrt{m}}{\xi'}\right),\]
which concludes the proof. \hfill$\blacksquare$
\section{Conclusion}\label{section6}
Here we propose a method to analyze the \textcolor{black}{dimensionality-dependent generalization bounds} for $k$-dimensional coding schemes, which are the abstract and general descriptions of a set of methods that encode random vectors in Hilbert space $\mathcal{H}$. There are several specific forms of $k$-dimensional coding schemes, including NMF, dictionary learning, sparse coding, $k$-means clustering and vector quantization, which have achieved great successes in pattern recognition and machine learning.

Our proof approach is based on an upper bound for the covering number of the loss function class induced by the reconstruction error. We explained that the covering number is more suitable for deriving dimensionality-dependent generalization bounds for $k$-dimensional coding schemes, because it avoids the worst case dependency \emph{w.r.t.} the number $k$ of the columns of the linear implementation. If $k$ is larger than the dimensionality $m$, our bound could be much tighter than the dimensionality-independent generalization bound. Moreover, according to Bennett's inequality, we derived a dimensionality-dependent generalization bound of order $\mathcal{O}\left(mk\ln(mkn)/n\right)^{\lambda_n}$, where $\lambda_n>0.5$ when the sample size $n$ is finite, for $k$-dimensional coding schemes. Our method therefore provides state-of-the-art dimensionality-dependent generalization bounds for NMF, dictionary learning, sparse coding, $k$-means clustering and vector quantization.

\bibliographystyle{apacite}
\bibliography{nmf}
\end{document}